\documentclass[journal]{IEEEtran}

\usepackage{threeparttable}
\usepackage{subfigure}
\usepackage{color}
\usepackage{multirow}
\usepackage{booktabs}
\usepackage{amssymb}

\usepackage{amsthm}
\usepackage{utfsym}

\usepackage{cite}

\usepackage{graphicx}
\ifCLASSINFOpdf
\else
\fi
\usepackage{amssymb}
\usepackage{amsmath}
\usepackage[capitalize,noabbrev]{cleveref}

\usepackage{algorithm}
\usepackage{algpseudocode}

\begin{document}
%

\title{Diffusion Disambiguation Models for Partial Label Learning}
\author{Jinfu~Fan,
	   Xiaohui~Zhong,
	   Kangrui~Ren,
	   Jiangnan~Li,
	    Linqing~Huang,~\IEEEmembership{Member,~IEEE,}   
	    Min~Gan,~\IEEEmembership{Senior Member,~IEEE,}   
	    C.~L.~Philip~Chen, ~\IEEEmembership{Fellow,~IEEE}   
\IEEEcompsocitemizethanks{\IEEEcompsocthanksitem 
Jinfu Fan, Xiaohui Zhong, Jiangnan Li, Min Gan and C. L. Philip Chen are with the College of Computer Science and Technology, Qingdao University, Qingdao 266071, China. (E-mail: fan\_jinfu@163.com, zhongxiaohui@qdu.edu.cn, lijiangnan@qdu.edu.cn, aganminn@aliyun.com and philip.chen@ieee.org). Kangrui Ren is with the Department of Control Science and Engineering, College of Electronics and Information Engineering, Tongji University, Shanghai 201804, China. (E-mail: 2310213@tongji.edu.cn). Linqing Huang is with the School of Electronic Information and Electrical Engineering, Shanghai Jiao Tong University, Shanghai 200240, China. (E-mail: huanglinqing95@gmail.com). Linqing Huang and Min Gan are the corresponding authors.}
}


\markboth{IEEE Transactions on Circuits and Systems for Video Technology}%
{Diffusion Disambiguation Models for Partial Label Learning}

\IEEEpubid{\parbox{\textwidth}{\centering Copyright © 2026 IEEE. Personal use of this material is permitted. \\ 
However, permission to use this material for any other purposes must be obtained from the IEEE by sending an email to pubs-permissions@ieee.org.}}

\maketitle

\begin{abstract}

 Learning from ambiguous labels is a long-standing problem in practical machine learning applications. The purpose of \emph{partial label learning} (PLL) is to identify the ground-truth label from a set of candidate labels associated with a given instance. Inspired by the remarkable performance of diffusion models in various generation tasks, this paper explores their potential to denoise ambiguous labels through the reverse denoising process. Therefore, this paper reformulates the label disambiguation problem from the perspective of generative models, where labels are generated by iteratively refining initial random guesses. This perspective enables the diffusion model to learn how label information is generated stochastically. By modeling the generation uncertainty, we can use the maximum likelihood estimate of the label for classification inference. However, such ambiguous labels lead to a mismatch between instance and label, which reduces the quality of generated data. To address this issue, this paper proposes a \emph{diffusion disambiguation model for PLL} (DDMP), which first uses the potential complementary information between instances and labels to construct pseudo-clean labels for initial diffusion training. Furthermore, a transition-aware matrix is introduced to estimate the potential ground-truth labels, which are dynamically updated during the diffusion generation. During training, the ground-truth label is progressively refined, improving the classifier. Experiments show the advantage of the DDMP and its suitability for PLL.

\end{abstract}


\IEEEpeerreviewmaketitle

\section{Introduction}

\begin{figure}[tbp] 
	\centering
	\includegraphics[width=0.86\linewidth]{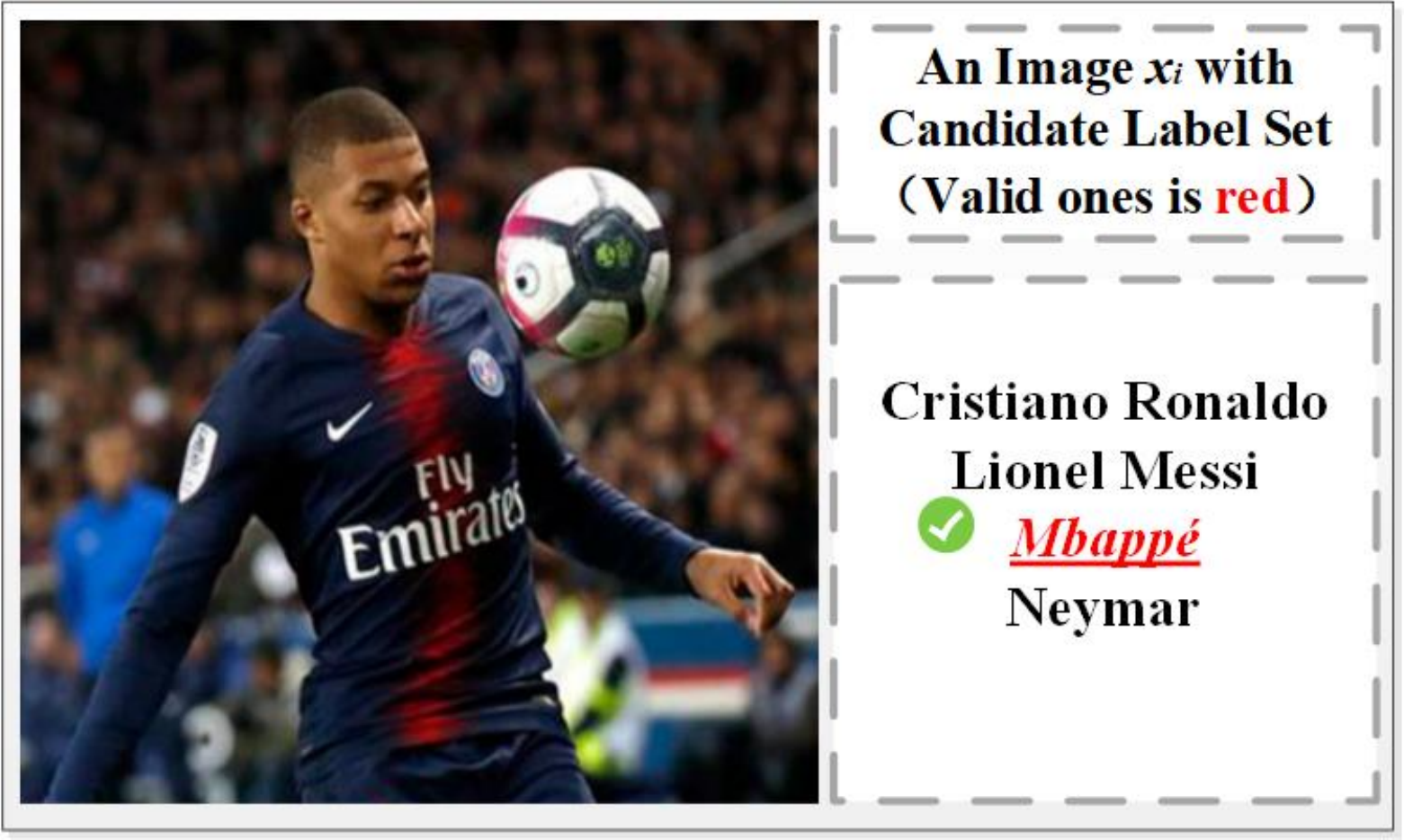}
	\caption{An example of partial label learning with noisy labels.}
	\label{fig1}
\end{figure}

\IEEEPARstart{S}{u}pervised learning has achieved remarkable accuracy in various classification tasks, but it typically requires large amounts of labeled data. Nonetheless, large-scale data annotation is often costly and labor-intensive, yet the labeled data often suffers from inherent inaccuracies stemming from either human mistakes or limitations in algorithmic labeling approaches. Additionally, research has shown that learning with ambiguous labels can cause models to overfit the noisy labels, reducing their generalization ability \cite{chen2024unm}. Therefore, it is essential to leverage \emph{partial label learning} (PLL) algorithms to effectively utilize cheap yet imperfect datasets for supervised learning tasks \cite{wang2022partial}. PLL is a form of weakly supervised learning in which each training instance is associated with a set of candidate labels, but only one of them represents the true label. PLL arises in many real-world scenarios such as image annotation (as shown in \cref{fig1}), object detection, web mining, etc \cite{xu2022multi}\cite{6381412}\cite{lu2025partial}. In such scenarios, the primary difficulty lies in accurately pinpointing the ground-truth label from the set of candidate labels. In this respect, related studies on noisy-label learning suggest that progressively correcting unreliable supervision can improve training robustness \cite{wang2022reflective}. However, PLL is fundamentally different from noisy-label learning, because it requires selecting the true label from multiple candidate labels rather than correcting a single corrupted annotation.

\IEEEpubidadjcol

To mitigate the impact of ambiguous labels, two general machine learning strategies have been proposed, i.e. \emph{non-deep learning strategy} (NDLS) \cite{berg2004s}\cite{luo2010learning}\cite{chai2019large} and \emph{deep learning strategy} (DLS) \cite{fan2026evidential}\cite{yao2021network}\cite{fan2021partial}. NDLS employs various optimization techniques to iteratively refine label confidence, aiming to identify the ground-truth labels from the candidate label set. In contrast, DLS incorporates some training objectives compatible with stochastic optimization, such as adding regularization terms, data augmentation, and adjusting loss weight. However, the drawback of the above methods is that they overly rely on observations of similar instances to estimate the uncertainty of labels, which can lead to incorrect disambiguation of uncertain labels in negative nearest neighbor instances. Negative nearest neighbor usually refers to the nearest neighbor of the opposite class to the current instance, which includes instances with similar features but not the same labeled class among the $k$-nearest neighbors calculated by instance features. Such inaccuracies progressively build up and persistently influence the learning process of the model, ultimately constraining its effectiveness in partial label learning scenarios.

To tackle this issue, a novel \emph{diffusion disambiguation model for PLL} (DDMP) is proposed. To alleviate the incorrect disambiguation caused by reliance on instance representation, we disambiguate candidate labels by modeling the generation process of ambiguous labels, which provides a new perspective on solving the PLL problem.  Inspired by the remarkable performance of diffusion models \cite{ho2020denoising} in various generation tasks, this paper explores their potential to denoise ambiguous labels through the reverse denoising process. But to the best of our knowledge, there is no theoretical discussion and research on the diffusion model in PLL learning. Recently,  \emph{Classification and Regression Diffusion Models} (CARD)\cite{han2022card} have extended diffusion models to address classification and regression problems. Inspired by CARD, DDMP addresses ambiguous label learning by treating it as a random process where labels are conditionally generated via diffusion mechanisms. The challenge of ambiguous label learning is approached as a stochastic label diffusion process conditioned on instance features. DDMP consists of a forward process and a reverse process. In the forward process, we start to make noisy estimates of the ambiguous labels and then gradually reconstruct them to learn to recover the ground-truth labels as output, which is essentially the same as the reverse denoising step in a diffusion model. The advantage of DDMP is that it introduces a diffusion model to gradually learn diffusion information in the feature space, in order to better simulate the relationships and uncertainty propagation between labels, and achieve gradual disambiguation from candidate sets to real labels. This is because the diffusion disambiguation in DDMP avoids the instability of traditional algorithms that rely solely on single predictions. Through iterative diffusion, DDMP aggregates global information, thereby mitigating the impact of erroneous candidate labels. \emph{\textbf{This task presents a significant challenge due to the presence of ambiguous labels. When training a diffusion model on a dataset with ambiguous labels, the model learns the conditional score associated with noisy labels.}}

To reduce the ambiguity of candidate labels, a smoothness assumption can be adopted, that is similar instances are more likely to have the same label. To quantify label similarity, the Jaccard distance is used to compute a label similarity matrix, enabling deeper exploration of relationships between candidate labels. For instance relations, especially negative nearest neighbor instances, the label adjacency matrix can mitigate the impact of similar instances with different labels and avoid the output of an instance being overwhelmed by the output of its negative nearest instance. Therefore, we leverage the potential complementary information between instances and labels to construct pseudo-clean labels that can preliminarily alleviate the impact of ambiguous labels during diffusion model training. Furthermore, a transition-aware matrix is introduced to estimate the potential ground-truth labels, which are dynamically updated during the diffusion generation. 
Meanwhile, we theoretically prove the mutual enhancement between the pseudo-clean label matrix and the transition-aware matrix from the perspective of the \emph{Expectation-Maximization} (EM) algorithm. In addition, as shown in \cref{fig2}, this paper designs a diffusion model structure suitable for PLL, which instance guided cross attention and label guided cross attention are beneficial for the model to fully explore the potential effective information between instances and labels, and perform feature fusion to guide the model in learning how to denoise. Meanwhile, DDMP includes a flexible conditional injection mechanism that supports time steps to guide model learning, enabling the model to effectively learn complex functions that map from noisy labels to disambiguation labels. The forward denoising and reverse disambiguation processes iteratively improve performance, progressively refining ground-truth labels. These refined labels in turn help to enhance classifier training. The advantage of this approach is that the probability of noise labels in the candidate label set is gradually reduced during back-diffusion, while the true label is amplified through global context and instance conditional information. Compared to a one-step label disambiguation strategy,  the multi-step denoising process can more smoothly correct labels and reduce the risk of over-reliance on local erroneous neighbors. In summary, the contributions of this paper are as follows:

\begin{itemize}  
	\item A novel \emph{diffusion disambiguation model for PLL} (DDMP) is proposed, which reframes learning from ambiguous labels as a stochastic process of conditional label generation, enabling effective learning from data with ambiguous labels.
	\item 
	The pseudo-clean label matrix is redesigned based on information complementarity, leveraging the potential interactions between instances and labels to prevent the output of an instance from being overwhelmed by its nearest negative instance.
	\item An adaptive estimator strategy with a transition-aware matrix is proposed to correct ambiguous labels, with a theoretical foundation based on the EM perspective.
\end{itemize}


Comprehensive experiments validate the robustness and effectiveness of DDMP, showing its performance surpasses that of state-of-the-art methods on partial label datasets.

\section{Related Work} \label{section_2}

\subsection{Partial Label Learning}

Existing PLL algorithms fall into two main categories, i.e., \emph{deep learning strategy} (DLS) and \emph{non-deep learning strategy} (NDLS).


\subsubsection{Non-deep Learning Strategy} \label{2.1.1}

\ 
\newline
\indent NDLS is built on the assumption that similar instances are likely to share the same label, and it resolves label ambiguity by exploring the underlying relationships between the candidate labels of similar instances. Various machine learning techniques, including \emph{Expectation Maximization} (EM) \cite{berg2004s}\cite{liu2012conditional}, \emph{K-Nearest Neighbors} (KNN)\cite{luo2010learning}\cite{zhang2016partial}, \emph{Support Vector Machine} (SVM) \cite{chai2019large} \cite{fan2023multi} and \emph{Low-Rank Representation} (LRR) \cite{fan2022addressing}\cite{chen2013dictionary}, have been widely applied to address PLL challenges. In addition, Zhang et al. \cite{chai2019large} further deepened the mining of inter-label relationships by constructing a weighted graph and utilizing affinity analysis to achieve iterative propagation of label information and effective differentiation of ground-truth label. Wang et al. \cite{wang2022partial} introduced a discrimination enhancement strategy for PLL to improve label confidence by reinforcing discriminative features in the representation space.

\begin{figure*}[t] 
	\centering
	\includegraphics[width=0.95\linewidth]{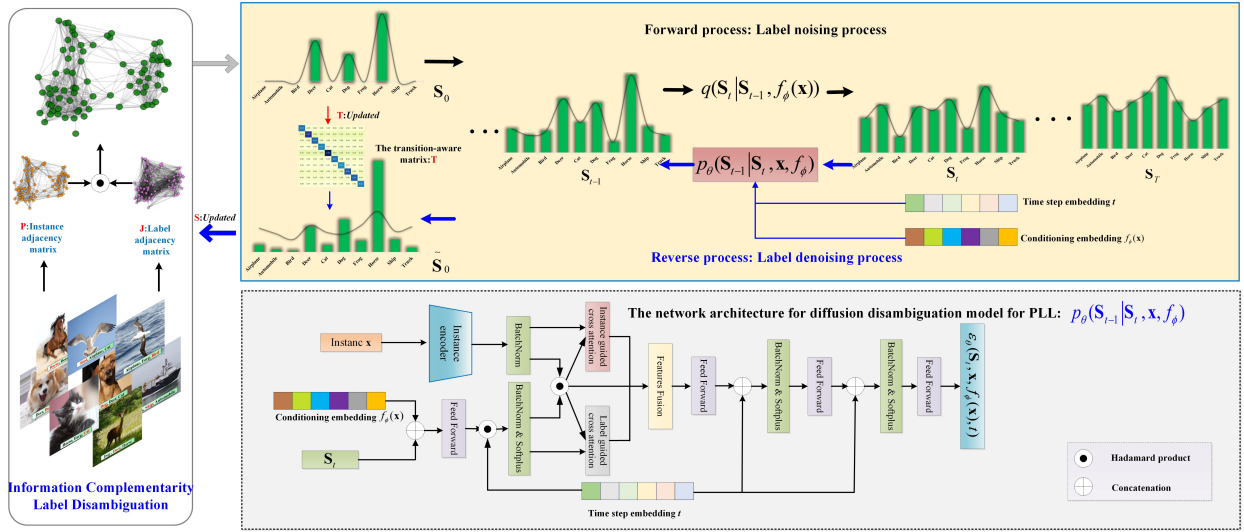}
	\caption{The training procedure of the proposed DDMP approach.}
	\label{fig2}
\end{figure*}

However, the time complexity of most NDLS algorithms grows significantly with larger datasets. To address this limitation, an increasing number of researchers are turning to deep learning approaches to tackle the PLL problem.

\subsubsection{Deep Learning Strategy} \label{2.1.2}
\ 
\newline
\indent The advancement of deep neural networks has spurred extensive research into establishing a deep learning frameworks for PLL. Yao et al. \cite{yao2020deep} pioneered the integration of PLL with deep learning by employing temporal coding techniques to combine outputs across different periods, guiding subsequent predictions. In the same year, Yao et al. \cite{yao2021network} enhanced label disambiguation in network interactions through a cooperative network mechanism. Lv et al. \cite{lv2020progressive} introduced a progressive recognition algorithm that iteratively refines ground-truth labels as the model evolves, ensuring strong compatibility. Wen et al. \cite{wen2021leveraged} incorporated a leverage parameter into the loss function, demonstrating promising experimental results across various deep learning frameworks. Fan et al. \cite{fan2021partial} proposed a competitive disambiguation strategy to reduce the influence of label ambiguity by updating the cluster center through competitive weighted disambiguation of neighboring instances. Besides, Fan et al. \cite{fan2023graphdpi} integrated graph representation learning into the PLL problem, developing a disambiguation correction network that significantly improved model performance in complex PLL scenarios by iteratively refining the disambiguation and correction processes. Subsequently, Fan et al. \cite{fan2022partial} further improved graph-based PLL by incorporating mutual information maximization for deeper partial label disambiguation. More recently, Wang et al. \cite{wang2023pico+} introduced contrastive learning to PLL, leveraging the Momentum Contrast strategy to greatly enhance the representation learning ability of the model. Feng et al. \cite{feng2025partial} studied PLL from the perspective of mutual information representation learning, improving partial-label disambiguation by enhancing feature learning and strengthening the consistency between learned representations and ambiguous supervision. Additionally, Fan et al.\cite{fan2024kmt} proposed a K-means cross-attention transformer for PLL, enabling continuous learning of cluster centers for effective disambiguation tasks. Realistically, instance-dependent partial label learning emerges as a more realistic framework for handling ambiguous supervision scenarios. To address this challenge,  a series of methods like contrastive learning, Bayesian models and variational inference have been proposed for instance dependent partial label learning \cite{xu2021instance, xu2024variational, xia2022ambiguity, wu2024distilling}.

One possible drawback of DLS is that it depends heavily on the relationships among candidate labels of similar instances, making it vulnerable to the influence of misleading or negative neighbor instances. To address this limitation, DDMP reformulates PLL as a stochastic conditional label generation and reverse denoising process. By combining information-complementarity-based pseudo-clean label construction, diffusion-based progressive disambiguation in label space, and dynamic transition-aware refinement, DDMP provides a new generative framework for more stable and robust partial-label disambiguation.


\subsection{Diffusion Model}

Generative models represent a crucial category of models in the realm of computer vision, and the diffusion model (DDPM) \cite{ho2020denoising} \cite{cao2025generative} is very popular nowadays compared to VAE \cite{kingma2013auto} and GAN \cite{goodfellow2020generative}. Diffusion models are inspired by the diffusion phenomenon in physics, where particles migrate from areas of higher concentration to areas of lower concentration. In the context of machine learning, this idea is harnessed for data synthesis by first applying a forward process that incrementally corrupts data with noise, and then training a reverse process that learns to remove that noise to recreate clean samples. During training, the  objective of the model is to estimate the exact noise added at each forward step, and those estimates are then used in the reverse (denoising) process to generate new data. To speed up the sampling process, DDIM \cite{song2020denoising} has emerged, which can be accelerated by skipping steps. CARD \cite{han2022card} extends the diffusion model to the domain of classification and regression, which can be used to solve the labeling problem.

However, when the diffusion model faces ambiguous labeling, it presents a significant challenge for the reverse denoising process. During training on datasets with ambiguous labels, the model learns the conditional scores associated with noisy labels, which significantly reduces the performance of the model.

\section{DDMP Method}
\subsection{Problem Formulation}

This section formulates the problem of PLL. We define the feature space as $\mathcal{X}\in\mathbb{R}^{d}$ and the output space as $\mathcal{Y}=\left\{  y_{c}:c=1,\ldots,Q \right\}$, where $d$ is the feature dimension and $Q$ is the number of classes. The training set for PLL is $\mathcal{D}=\left\{ (\mathbf{x}_{i},S_{i}) |1\leq i\leq N \right\}$, where $\mathbf{x}_{i}\in \mathcal{X}$ represents instance features and $S_{i}\subseteq \mathcal{Y}$ represents the candidate label set associated with $\mathbf{x}_{i}$. Particularly, $\begin{vmatrix} S_{i} \end{vmatrix}$ represents the number of ambiguous labels for instance $\mathbf{x}_{i}$. Meanwhile, the label confidence matrix can be denoted as  $\mathbf{Y}=\left[ \mathbf{y}_{1},\mathbf{y}_{2},\ldots,\mathbf{y}_{N} \right] ^{T}\in\mathbb{R}^{N \times Q}$, where each $\mathbf{y}_{i} \in\{0,1\}^{Q}$ marks its candidate labels with $1$ (and non‐candidates with $0$). The aim of DDMP is then to learn a classifier $\boldsymbol{f}:\mathcal{X}\mapsto \mathcal{Y}$ that can correctly predict labels for previously unseen inputs.


\subsection{The Forward Process of DDMP}

Much like in traditional diffusion frameworks \cite{ho2020denoising,chen2024label}, DDMP is built around a forward and a reverse pass. During forward diffusion, the $Q$-dimensional  pseudo-clean label $\mathbf{S}_{0}$ undergoes progressive corruption across $T$ timesteps, yielding intermediate states $\mathbf{S}_{1:T}$. This converges asymptotically to  $\mathcal{N}\left(f_{\phi}(\mathbf{x}), \mathbf{I}\right)$ (latent distribution), where the mean $f_{\phi}$ comes from a pre-trained instance encoder, encoding prior information about the relationship between $\mathbf{x}$ and $\mathbf{S}_{0}$. The initial pseudo-clean label $\mathbf{S}$ is calculated based on complementary information between instances and candidate labels, and is continuously updated during subsequent iterations, which will be described in detail later. The forward process in diffusion models, denoted as $q(\mathbf{S}_{1:T} | \mathbf{S}_{0},f_{\phi}(\mathbf{x}))$, is also a Markov chain. It is defined by adding Gaussian noise to the data over time, and the form of the forward process conditional distributions in a similar fashion as\cite{pandey2022diffusevae}, which can be summarized as follows:



\begin{equation} \label{1}
	\begin{aligned}
		q(\mathbf{S}_{1:T} | \mathbf{S}_{0},f_{\phi}(\mathbf{x})) = \prod_{t=1}^{T} q(\mathbf{S}_{t} | \mathbf{S}_{t-1},f_{\phi}(\mathbf{x}))
	\end{aligned}
\end{equation}

\begin{equation} \label{2}
	\begin{aligned}
		&q\left(\mathbf{S}_{t} \mid \mathbf{S}_{t-1}, f_{\phi}(\mathbf{x})\right)= \\
		&\mathcal{N}\left(\mathbf{S}_{t} ; \sqrt{1-\beta_{t}} \mathbf{S}_{t-1}+\left(1-\sqrt{1-\beta_{t}}\right) f_{\phi}(\mathbf{x}), \beta_{t} \mathbf{I}\right)
	\end{aligned}
\end{equation}

\noindent where $\left\{ \beta_{t} \right\} _{t=1:T} \in(0,1)^{T}$ is the variance schedule. This recursive formulation allows direct sampling of $\mathbf{S}_{t}$, with $t$ an arbitrary timestep,


\begin{equation} \label{3}
	\begin{aligned}
		&q\left(\mathbf{S}_{t} \mid \mathbf{S}_{0}, f_{\phi}(\boldsymbol{x})\right)= \\
		&\mathcal{N}\left(\mathbf{S}_{t} ; \sqrt{\bar{\alpha}_{t}} \mathbf{S}_{0}+\left(1-\sqrt{\bar{\alpha}_{t}}\right) f_{\phi}(\mathbf{x}), (1-\bar{\alpha}_{t}) \mathbf{I}\right)
	\end{aligned}
\end{equation}

\noindent where $\alpha_{t}=1-\beta_{t}$ and $\bar{\alpha}_{t} = \prod_{t} \alpha_{t}$, which shows that we can sample any noisy version $\mathbf{S}_{t}$ within a single step, with a fixed variance $\beta_{t}$. Sampling from $q\left(\mathbf{S}_{t} \mid \mathbf{S}_{0}, f_{\phi}(\boldsymbol{x})\right)$ is performed via the reparametrization trick. From the above equation, the mean term can be regarded as an interpolation between the initial candidate label $\mathbf{S}_{0}$ and the mean of the latent distribution $f_{\phi}(\mathbf{x})$, gradually changing from the former to the latter throughout the entire forward process. Consequently, $\mathbf{S}_{t}$ can be computed as:

\begin{equation} \label{4}
	\begin{aligned}
		\mathbf{S}_{t} = \sqrt{\bar{\alpha}_{t}}\mathbf{S}_{0} + (1 - \sqrt{\bar{\alpha}_{t}})f_{\phi}(\mathbf{x}) + \sqrt{1 - \bar{\alpha}_{t}}\boldsymbol{\epsilon},
	\end{aligned}
\end{equation}

\noindent where $\boldsymbol{\epsilon}\sim  \mathcal{N}\left(\mathbf{0}, \mathbf{I}\right)$. Detailed derivations can be found in \cref{Proof1} .

\textbf{Information Complementarity Label Disambiguation Strategy for Training:} When training a diffusion model on a dataset with ambiguous labels, the model learns the conditional score associated with noisy labels. To mitigate this issue, we propose a strategy that leverages complementary information from instances and candidate labels to identify potential ground-truth labels, making the model more resistant to ambiguous labels. Our main assumption is that in the latent space, data points from the same class have the same manifold structure. Therefore, most of the neighbors of the data point should have the same label as the point itself. To this end, we adopt the $k$-nearest neighbors of the instance embedding features, calculated as a pre-trained encoder $f_{\phi}$, to construct the instance adjacency matrix $\mathbf{P}$, where $\mathbf{P}_{ij}=1$ if $f_{\phi}(\mathbf{x}_{i}) \in \mathcal{N}_{k}(f_{\phi}(\mathbf{x}_{j}))$ or $f_{\phi}(\mathbf{x}_{j}) \in \mathcal{N}_{k}(f_{\phi}(\mathbf{x}_{i}))$, otherwise $\mathbf{P}_{ij}=0$. The $\mathcal{N}_{k}(f_{\phi}(\mathbf{x}_{j}))$ is deﬁned as the $k$-nearest neighbor set of $f_{\phi}(\mathbf{x}_{j})$. Furthermore, the candidate labels provide potentially useful information to reduce label ambiguity. For example, if two instances have no overlapping candidate labels, they remain unlinked. Otherwise, a link is established between them. Hence, following Eq.($\ref{5}$), we use the Jaccard distance to measure the similarity between their sets of candidate labels.


\begin{equation} \label{5}
	\begin{split}
		\mathbf{J}_{ij}(S_{i},S_{j})&=1- \frac{\begin{vmatrix} S_{i}\cup S_{j} \end{vmatrix} - \begin{vmatrix} S_{i} \cap S_{j} \end{vmatrix}}{\begin{vmatrix} S_{i}\cup S_{j} \end{vmatrix} } \in [0,1]
	\end{split}
\end{equation}

\noindent where $\mathbf{J} \in \mathbb{R}^{N \times N}$ is label similarity matrix. Therefore, we can perform disambiguation on candidate labels based on the complementary information of $\mathbf{P}$ and $\mathbf{J}$. For instance, even when two instances are similar ($\mathbf{P}_{ij}>0$), they won’t be linked if they have no shared labels ($\mathbf{J}_{ij}=0$), because ($\mathbf{P}_{ij}\times \mathbf{J}_{ij}=0$). Likewise, if they share labels ($\mathbf{J}_{ij}>0$) but are not similar ($\mathbf{P}_{ij}=0$), there will still be no connection. Therefore, complementary information between instances and candidate labels can be used to improve the accuracy of disambiguation. The initial pseudo-clean label matrix $\mathbf{S}\in \mathbb{R}^{N \times Q}$ can be calculated by $\mathbf{S} = (\mathbf{P} \odot \mathbf{J})\mathbf{Y}$, where $\odot$ means hadamard product. Then, DDMP is trained to learn the conditional distribution $p(\mathbf{S}|\mathbf{x})$ of disambiguated labels, rather than the label distribution of the data point itself.


Information complementarity label disambiguation strategy enables models to utilize information from multiple potentially more accurate labels to improve their predictive performance. In addition, as the diffusion model is effective in modeling multimodal distributions, training the model to mine potential information in the data to generate corresponding labels can reduce the inherent uncertainty in the data annotation process\cite{chen2024label}. Therefore, in order to further improve the effectiveness of the model, the pseudo-clean labels in this paper are not fixed but are dynamically updated. This can gradually improve the correspondence between instances and labels, thereby enhancing model performance.

\subsection{The Reverse Process of DDMP}

The reverse pass can likewise be modeled as a first-order Markov chain, where the transition dynamics are governed by a learned Gaussian distribution, we start a $Q$-dimensional Gaussian noise from $\mathbf{S}_{T} \sim  \mathcal{N}\left(f_{\phi}(\mathbf{x}), \mathbf{I}\right)$ and reverse the process using $p_{\theta}\left(\mathbf{S}_{t-1} \mid \mathbf{S}_{t}, \mathbf{x}, f_{\phi}(\mathbf{x})\right)$ to reconstruct a label vector $\mathbf{S}_{0}$.


\begin{equation} \label{6}
	\begin{aligned}
		p(\mathbf{S}_{0:T} | \mathbf{x},f_{\phi}(\mathbf{x})) = p(\mathbf{S}_{T})\prod_{t=1}^{T} p_{\theta}(\mathbf{S}_{t-1} | \mathbf{S}_{t},\mathbf{x}, f_{\phi})
	\end{aligned}
\end{equation}

\begin{equation} \label{7}
	\begin{aligned}
		p_{\theta}\left(\mathbf{S}_{t-1} \mid \mathbf{S}_{t}, \mathbf{x}, f_{\phi}\right)=\mathcal{N}\left(\mathbf{S}_{t-1} ; \boldsymbol{\mu}_{\theta}\left(\mathbf{S}_{t}, \mathbf{x}, f_{\phi}, t\right), \tilde{\beta}_{t} \mathbf{I}\right)
	\end{aligned}
\end{equation}

\noindent where $\tilde{\beta}_{t}=\frac{1-\bar{\alpha}_{t-1} }{1-\bar{\alpha}_{t} }\beta_{t}$, the transition step is Gaussian for an infinitesimal variance $\beta_{t}$. The diffusion model is learned through a neural network to emulate $p_{\theta}\left(\mathbf{S}_{t-1} \mid \mathbf{S}_{t}, \mathbf{x}, f_{\phi}(\mathbf{x})\right)$ by optimizing the evidence lower bound with stochastic gradient descent:

\begin{equation} \label{8}
	\begin{aligned}
		\mathcal{L}&=\mathbb{E}_{q}\left[\mathcal{L}_{T}+\sum_{t>1}^{T} \mathcal{L}_{t-1}+\mathcal{L}_{0}\right], \text { where }\\
		\mathcal{L}_{0}&=-\log p_{\theta}\left(\mathbf{S}_{0} \mid \mathbf{S}_{1}, \mathbf{x}, f_{\phi}\right) \\
		\mathcal{L}_{t-1}&=D_{\mathrm{KL}}\left(q\left(\mathbf{S}_{t-1} \mid \mathbf{S}_{t}, \mathbf{S}_{0}, \mathbf{x}, f_{\phi}\right) \| p_{\theta}\left(\mathbf{S}_{t-1} \mid \mathbf{S}_{t}, \mathbf{x}, f_{\phi}\right)\right) \\
		\mathcal{L}_{T}&=D_{\mathrm{KL}}\left(q\left(\mathbf{S}_{T} \mid \mathbf{S}_{0}, \mathbf{x}, f_{\phi}\right)| | p\left(\mathbf{S}_{T} \mid \mathbf{x}, f_{\phi}\right)\right)
	\end{aligned}
\end{equation}

Eq.($\ref{8}$) focuses on  the closeness of $p_{\theta}\left(\mathbf{S}_{t-1} \mid \mathbf{S}_{t}, \mathbf{x}, f_{\phi}(\mathbf{x})\right)$ to the true posterior of the forward process. Following\cite{han2022card}, the mean term is written as $\boldsymbol{\mu}_{\theta}\left(\mathbf{S}_{t}, \mathbf{x}, f_{\phi}, t\right) = \frac{1}{\alpha_{t}}(\mathbf{S}_{0}-\frac{\beta_{t}}{\sqrt{1 - \bar{\alpha}_{t}}}\mathbf{\epsilon}_{\theta}(\mathbf{S}_{t},\mathbf{x},f_{\phi}, t))$ and the objective can be simplified to $\mathcal{L}=\left\|\boldsymbol{\epsilon}-\boldsymbol{\epsilon}_{\theta}\left(\mathbf{S}_{t}, \mathbf{x}, f_{\phi}, t\right)\right\|^{2}$. Then, we predict the ambiguous label $\tilde{\mathbf{S}}_{0}$, a prediction of $\mathbf{S}_{0}$ given $\mathbf{S}_{t}$ as:

\begin{equation} \label{9}
	\begin{aligned}
		&\tilde{\mathbf{S}}_{0} = 
		\frac{1}{\sqrt{\alpha_{t}}} \left[ \mathbf{S}_{t} - (1 - \sqrt{\overline{\alpha}_{t}}) f_{\phi} - \sqrt{1 - \overline{\alpha}_{t}} \, \boldsymbol{\epsilon}_{\theta}(\mathbf{S}_{t}, \mathbf{x}, f_{\phi}, t) \right]
	\end{aligned}
\end{equation}

%
%
%
%

\textbf{Ambiguous Labels:} For PLL, each partially labeled instance is independently drawn from a probability distribution, and the key assumption is that correct label $y$ of $\mathbf{x}$ must be in the candidate label set. Therefore, from a generative perspective, it can be expressed as follows:

\begin{equation} \label{10}
	\begin{aligned}
		p(S|\mathbf{x}) &= \sum_{c=1}^{Q}p(S|y=y_{c})p(y = y_{c}|\mathbf{x})
	\end{aligned}
\end{equation}

The proof of Eq.($\ref{10}$) is provided in \cref{Proof3}, which means that the clean class-posterior $p(y = y_{c}|\mathbf{x})$ can be inferred by utilizing the ambiguous class-posterior $p(S|\mathbf{x})$. This formulation relies on the assumption \(p(S|\mathbf{x},y)=p(S|y)\), which is consistent with the standard synthetic PLL protocol adopted in this paper, where candidate labels are generated in an instance-independent manner conditioned on the ground-truth label. In real-world partially labeled scenarios, however, this assumption should be regarded as a modeling approximation rather than an exact description of the annotation process. Under this formulation, we introduce a transition-aware matrix $\mathbf{T}=[T_{ij}]^{Q \times Q}$, where $\mathbf{T}_{ij}=p( y_{i} \in S|y = y_{j})$, i.e., $p(y|\mathbf{x}) =[\mathbf{T}]^{-1}p(S|\mathbf{x})$. Then, the difficulty is how to estimate the transition-aware matrix.

%
%

\textbf{Estimation of Transition-Aware Matrix:} Generally, the transition-aware matrix is non-identifiable without any additional assumption. To implement the DDMP objective, we need to estimate the transition-aware matrix. Concretely, let $\hat{p}(S|\mathbf{x};\theta)$ parameterized by $\theta \in \mathcal{W}$ be a differentiable model for the ambiguous labels, and $\mathbf{T} \in \mathcal{T}$ be an estimator for the transition-aware matrix. Assuming that the function of $\hat{p}(S|\mathbf{x};\theta)$ is sufficiently large enough, this means that there exists an optimal parameter $\theta^{*} \in \mathcal{W}$ that makes $[\mathbf{T}]^{-1}\hat{p}(S|\mathbf{x};\theta^{*})$ and $p(y|\mathbf{x})$ equal everywhere. In practice, we use an expressive deep neural
network, the reverse model of DDMP, to approximate $p(S|\mathbf{x})$. 

We note that the above assumption is introduced to motivate the approximation capacity of the reverse model, rather than to claim strict identifiability or universal convergence of the transition-aware matrix in arbitrary real-world scenarios. In DDMP, the transition-aware matrix is treated as a practically motivated estimator that is iteratively refined together with the pseudo-clean label matrix during training. Therefore, the EM-based analysis should be understood as a theoretical interpretation of progressive ambiguity refinement, rather than a proof of exact recovery of the ground-truth transition structure under all conditions.

The transition-aware matrix is estimated by:


\begin{equation} \label{11}
	\begin{aligned}
		\mathbf{T}_{ij}^{e}=\frac{\sum_{k=1}^{N} \mathbb{I}(y_{j} \in S_{k})\mathbf{S}_{ki}^{e}}{\sum_{k=1}^{N} \mathbf{S}_{ki}^{e}}
	\end{aligned}
\end{equation}

\noindent where $\mathbb{I}(\cdot)$ is an indicator function. $e$ is the number of iterations. Since $\mathbf{S}$ is updated asymptotically, $\mathbf{T}$ is also dynamically estimated based on $\mathbf{S}$ to approximate the true value. We propose a softened and moving-average style strategy to update the ambiguous labels. The pseudo-clean labels are updated by:

\begin{equation} \label{12}
	\begin{aligned}
		\mathbf{S}^{e+1} = \textnormal{Normalize}((\mathbf{S}^{e}+[\mathbf{T}^{e}]^{-1}\tilde{\mathbf{S}}_{0}^{e})\mathbf{S}^{e})
	\end{aligned}
\end{equation}

\noindent where $\textnormal{Normalize}(\cdot)$ is normalization function. When $e = 1$, $\mathbf{S}^{1} =\mathbf{S}$ is initial pseudo-clean label matrix. Due to the fact that the parameters of the model are still far from optimal during early training, a moving‑average strategy is applied to gradually adjust the pseudo‑clean labels toward their true values while maintaining stable training dynamics. The k-NN graph and Jaccard-based label similarity are used as a stable initialization prior, rather than an explicitly adaptive neighbor reconstruction mechanism. The dynamic refinement in DDMP is mainly realized through pseudo-clean label updating, transition-aware estimation, and reverse denoising during training. More specifically, DDMP follows a two-level refinement mechanism. In the inner denoising process, a sampled diffusion timestep $t$ is used to generate the noisy label state and obtain the denoised label estimate $\tilde{\mathbf{S}}_{0}^{e}$. In the outer refinement process, the transition-aware matrix $\mathbf{T}^{e}$ is estimated under the current refinement state, and then $\mathbf{T}^{e}$ and $\tilde{\mathbf{S}}_{0}^{e}$ are jointly used to update the pseudo-clean label matrix from $\mathbf{S}^{e}$ to $\mathbf{S}^{e+1}$. Therefore, the interaction between label refinement and denoising progression is realized through an alternating outer-loop refinement process rather than a timestep-specific update of $\mathbf{T}$. As illustrated in \cref{fig2}, with the progression of iterative training, the ground-truth label is progressively identified through Eq.($\ref{12}$), and the refined labels can improve the information matching between the instances and labels. Therefore, the forward denoising and backward reverse disambiguation processes are alternately and iteratively improved in their performance. The training process is considered complete once both components achieve satisfactory performance. We further rigorously draw a resemblance of DDMP with a classical EM-style algorithm in \cref{Proof4}. \cref{alg:example1} summarizes the complete procedure of DDMP.

\begin{algorithm}[t]
	\caption{DDMP Framework}
	\label{alg:example1}
	\begin{algorithmic}
		\State {\bfseries Input:} Training set $\mathcal{D}$, label confidence matrix $\mathbf{Y}$, pre-trained encoder $f_{\phi}$.
		\State Initialization
		\State Calculate initial pseudo-clean matrix $\mathbf{S}$ according to $\mathbf{P}$ and $\mathbf{J}$;
		\While{not converged}
		\State Sample time slice $t \sim \left\{  1,\ldots,T \right\}$, data ($\mathbf{x}$,$\mathbf{S}$), and noise \Statex \qquad $\boldsymbol{\epsilon}\sim  \mathcal{N}\left(\mathbf{0}, \mathbf{I}\right)$
		\State Sample $\mathbf{S}_{t}$, and convert it to a one-hot vector $\mathbf{S}_{0}$
		\State Optimize the loss with a gradient descent step \Statex \qquad $\mathcal{L}=\left\|\boldsymbol{\epsilon}-\boldsymbol{\epsilon}_{\theta}\left(\mathbf{S}_{t}, \mathbf{x}, f_{\phi}, t\right)\right\|^{2}$
		\State Update transition-aware matrix $\mathbf{T}$ according to Eq.($\ref{11}$)
		\State Update pseudo-clean label matrix $\mathbf{S}$ according to Eq.($\ref{12}$)
		\EndWhile
	\end{algorithmic}
\end{algorithm}

\section{Experiments} \label{section_4}

The experimental results presented in this section demonstrate the superior performance of the proposed method on partial label datasets when compared to existing PLL techniques. Code and data can be found at: https://github.com/fanjinfucool/DDMP.


\subsection{Datasets} \label{4.1}

(1) \textbf{Real-World PLL datasets}.

To comprehensively and effectively evaluate the performance of the proposed DDMP method, we employ five widely used real-world partial label datasets, including Lost \cite{yao2018deep}, BirdSong \cite{briggs2012rank}, Soccer Player \cite{zeng2013learning}, MSRCv2 \cite{liu2012conditional}, and Yahoo! News \cite{guillaumin2010multiple}. The details of these datasets, including  \emph{number of features} ($\#$FES), \emph{number of instances} ($\#$INS), \emph{the mean number of ambiguous labels} ($\#$EAL), and \emph{number of labels} ($\#$LAS), are summarized in \cref{tab:5}. These datasets encompass diverse application scenarios, including \emph {Object Classification}, \emph {Automatic Face Naming} and \emph {Bird Song Classification}.


(2) \textbf{Synthesized PLL datasets}. 

To assess the effectiveness of the proposed algorithm, we utilize five commonly used benchmark datasets, including MNIST \cite{lecun1998gradient}, Kuzushiji-MNIST \cite{clanuwat2018deep}, CIFAR-10 \cite{krizhevsky2009learning}, Fashion-MNIST \cite{xiao2017fashion}, and CIFAR-100 \cite{krizhevsky2009learning}. The partial label datasets are constructed following the standard synthesis procedure in PLL \cite{yao2021network}, where candidate label sets for each instance are formed through $Q-1$ independent selections of noise labels. The parameter $q$ controls the probability of each noise label being included. Notably, higher values of $q$ increase the likelihood of instances having more noise labels, thereby raising the difficulty of label disambiguation. We experiment with $q$ values from the set $ \left\{  0.1,0.3,0.5 \right\}$ for MNIST, Kuzushiji-MNIST, Fashion-MNIST, CIFAR-10 and $ \left\{  0.01,0.05,0.1 \right\}$ for CIFAR-100.



\begin{table}[t]
	\centering
	\caption{Comprehensive Information on Real-World Datasets}	
	\label{tab:5}       
	\begin{tabular}{c cccc|c}
		\hline		
		\hline\noalign{\smallskip}		
		Data  &FEA  & INS & MAL & LAS & Scenarios  \\		
		\noalign{\smallskip}\hline\noalign{\smallskip}
		MSRCv2 & 48 & 1758  & 3.16 & 23 & \emph {Object Classification} \\
		Lost & 108 & 1122  & 2.33 & 16 & \emph {Automatic Face Naming} \\
		BirdSong & 38 & 4998  & 2.18 & 13 & \emph {Bird Song Classification} \\	
		Yahoo! News & 163 & 22991  & 1.91 & 219 & \emph {Automatic Face Naming} \\
		Soccer Player & 279 & 17472  & 2.09 & 171 & \emph {Automatic Face Naming} \\			
		\noalign{\smallskip}\hline
		\hline		
	\end{tabular}	
\end{table}

%
%
%
%
%

\begin{table*}[t]
	\scriptsize	
	\centering
	\caption{Accuracy comparisons on real-world partially labeled datasets.}
	\label{table2}
	\resizebox{\textwidth}{!}{
	\begin{tabular}{cccccccccccc}
		\hline
		\toprule
		\multirow{2}{*}{Datasets} & \multicolumn{11}{c}{Method} \\  
		\cmidrule(lr){2-12}  
		& PiCO   & CAVL &  VALEN & PRODEN & LW	& ABLE   & DIRK & CRDPLL & CroSel & LS-PLL & \textbf{DDMP}\\
		\midrule
		Lost      & 65.33 $\pm$ 0.75\%  &63.11 $\pm$ 0.72\% & 71.56 $\pm$ 0.80\% & 65.17 $\pm$ 0.72\%  & 67.11 $\pm$ 0.78\%  &68.93 $\pm$ 0.11\%  & 74.26 $\pm$ 0.58\% & 64.55 $\pm$ 0.31\% &72.89 $\pm$ 0.72\%& 66.67 $\pm$ 0.13\% & \textbf{74.58 $\pm$ 0.21\%}\\
		MSRCv2    & 49.14 $\pm$ 0.57\%  & 52.84 $\pm$ 0.16\% & 45.89 $\pm$ 0.55\% & 52.00 $\pm$ 0.46\% & 47.15 $\pm$ 0.75\% & 53.66 $\pm$ 0.95\%  & 44.61 $\pm$ 0.32\% &49.14 $\pm$ 0.87\%&48.58 $\pm$ 0.46\%&51.42 $\pm$ 0.38\% & \textbf{53.71 $\pm$ 0.28\%}\\
		BirdSong  & 61.29 $\pm$ 0.10\%  & 70.50 $\pm$ 0.92\% & 72.30 $\pm$ 0.49\% & 63.38 $\pm$ 0.52\% & 66.50 $\pm$ 0.73\% & 74.70 $\pm$ 0.78\%  & 71.28 $\pm$ 0.21\% & 72.00 $\pm$ 0.62\%&71.90 $\pm$ 0.89\%&56.80 $\pm$ 0.63\% & \textbf{78.96 $\pm$ 0.42}\%\\
		Soccer Player  & 55.13 $\pm$ 0.48\% & 54.27 $\pm$ 0.37\%  & 53.91 $\pm$ 0.12\% & 51.45 $\pm$ 0.19\% & 49.15 $\pm$ 0.38\%  & 62.77 $\pm$ 0.42\% & 53.37 $\pm$ 0.48\% & 54.47 $\pm$ 0.22\%&53.53 $\pm$ 0.10\%&54.65 $\pm$ 0.48\%& \textbf{62.83 $\pm$ 0.24\%}\\
		Yahoo! News & 68.71 $\pm$ 0.22\% & 63.86 $\pm$ 0.58\%  & 67.73 $\pm$ 0.32\% & 59.06 $\pm$ 0.38\% & 47.03 $\pm$ 0.37\% &53.35 $\pm$ 0.35\% & 61.38 $\pm$ 0.22\% & 65.23 $\pm$ 0.74\%&\textbf{69.38 $\pm$ 0.15}\%&62.56 $\pm$ 0.35\%& 54.56 $\pm$ 0.37\% \\
		
		\bottomrule
		\hline
	\end{tabular}}
\end{table*}

\subsection{Compared Methods} \label{4.2}

To demonstrate the superiority of the proposed DDMP algorithm, we conducted comparisons with state-of-the-art PLL methods, including deep learning based on PLL methods PRODEN \cite{lv2020progressive}, LW  \cite{wen2021leveraged}, CAVL \cite{zhang2021exploiting}, CRDPLL \cite{wu2022revisiting}, CroSel \cite{tian2024crosel}, LS-PLL \cite{gong2024does} and PiCO \cite{wang2023pico+}, and the
deep learning based on instance-dependent PLL methods VALEN \cite{xu2024variational}, ABLE \cite{xia2022ambiguity}, and DIRK \cite{wu2024distilling}. Parameter configurations for compared methods replicate literature specifications.



\begin{itemize}
	\item VALEN \cite{xu2024variational}: Predictive models are trained iteratively at
	each period using candidate labels and label distributions. [suggested configuration: Weight decay ($10^{-4}$), training batch($500$), learning rate ($10^{-2}$)].
	\item PRODEN \cite{lv2020progressive}: Model updating and label disambiguation are
	done step-by-step using an incremental recognition approach. [suggested configuration: Weight decay ($10^{-5}$), training batch($500$), learning rate ($10^{-3}$) ].
	\item LW  \cite{wen2021leveraged}: The trade-off between candidate label loss and
	non-candidate label loss is considered using leverage weighted loss. [suggested configuration: Weight decay ($10^{-5}$), training batch($500$), learning rate ($10^{-3}$) ].
	\item CAVL \cite{zhang2021exploiting}: Truth labels are identified using class activation graphs and their promotion form class activation values to improve partial label learning, and ultimately real labels are progressively identified through the intrinsic representation of the model. [suggested configuration: Learning rate ($10^{-3}$), training batch($250$), weight decay ($10^{-3}$) ].
	\item PICO \cite{wang2023pico+}: Utilizing the idea of learning by contrast for representation learning and ultimately disambiguation through prototype updating. [suggested configuration:  Learning rate ($10^{-2}$), training batch($500$), weight decay ($10^{-3}$) ].
	\item ABLE \cite{xia2022ambiguity}: Boosting PLL performance via contrastive learning explicitly leveraging label ambiguity. [suggested configuration: Learning rate ($10^{-2}$), training batch($500$), weight decay ($10^{-3}$) ].
	\item DIRK \cite{wu2024distilling}: A partial label learning algorithm synergizing knowledge distillation and contrastive learning to improve model robustness. [suggested configuration:  Learning rate ($10^{-2}$), training batch($500$), weight decay ($10^{-3}$) ].
	\item CRDPLL \cite{wu2022revisiting}: A PLL algorithm that enhances model performance through the use of consistency regularization. [suggested configuration: Training batch($64$), learning rate ($10^{-1}$), weight decay ($10^{-4}$), $T^{\prime}=100$, $\lambda=1$, and $K=3$].
	\item CroSel \cite{tian2024crosel}: Using dual model cross selection and consistency regularization to cross-select the most reliable pseudo labels from candidate labels for the other party, achieving high-precision recognition of real labels. [suggested configuration: Training batch($64$), learning rate ($10^{-1}$) ].
	\item LS-PLL \cite{gong2024does}: A baseline solution and a new optimization algorithm is designed called partial label learning based on label smoothing, which improves model performance by introducing label smoothing.[suggested configuration: Training batch($128$), learning rate ($10^{-2}$) ].
\end{itemize}

\subsection{Implementation Details} \label{4.3}

As shown in \cref{fig2}, the DDMP consists of a pre-trained instance encoder $f_{\phi}$, an instance encoder layer, an instance-guided cross attention, a label-guided cross attention, and a series of feed-forward layers. Two cross attention mechanisms are used to mine potential effective information of two encoding features through information guidance. These features are then combined with a Hadamard product and time embedding. A sequence of feed-forward networks predicts the noise term $\boldsymbol{\epsilon}_{\theta}$, utilizing batch normalization layers and Softplus activation functions. Our implementation is executed using PyTorch, and all experiments were conducted with NVIDIA GeForce RTX 4090 GPU. The DDMP framework is trained for 200 epochs with Adam optimization and 256 batch size. For the nearest neighbors, we set $k=10$. Following \cite{chen2024label}, the sampling trajectory is 10 and $T = 1000$. In our experiments, we use SimCLR \cite{chen2020simple} and CLIP \cite{radford2021learning} as pre-trained models respectively as shown in \cite{chen2024label}. For SimCLR, we use MLP for MNIST, Kuzushiji-MNIST, and Fashion-MNIST, and
ResNet34 for CIFAR-100 and CIFAR-10 as the pre-trained encoder. Since the real-world partially labeled datasets are processed, we directly use these data as encoding features. For CLIP, we utilized the Vision Transformer encoder (ViT-L/14), pre-trained within the CLIP framework, as it represents the top-performing architecture for this task.



\begin{figure}[t]
	\vskip 0in
	\begin{center}
		\centerline{\includegraphics[width=0.80\linewidth]{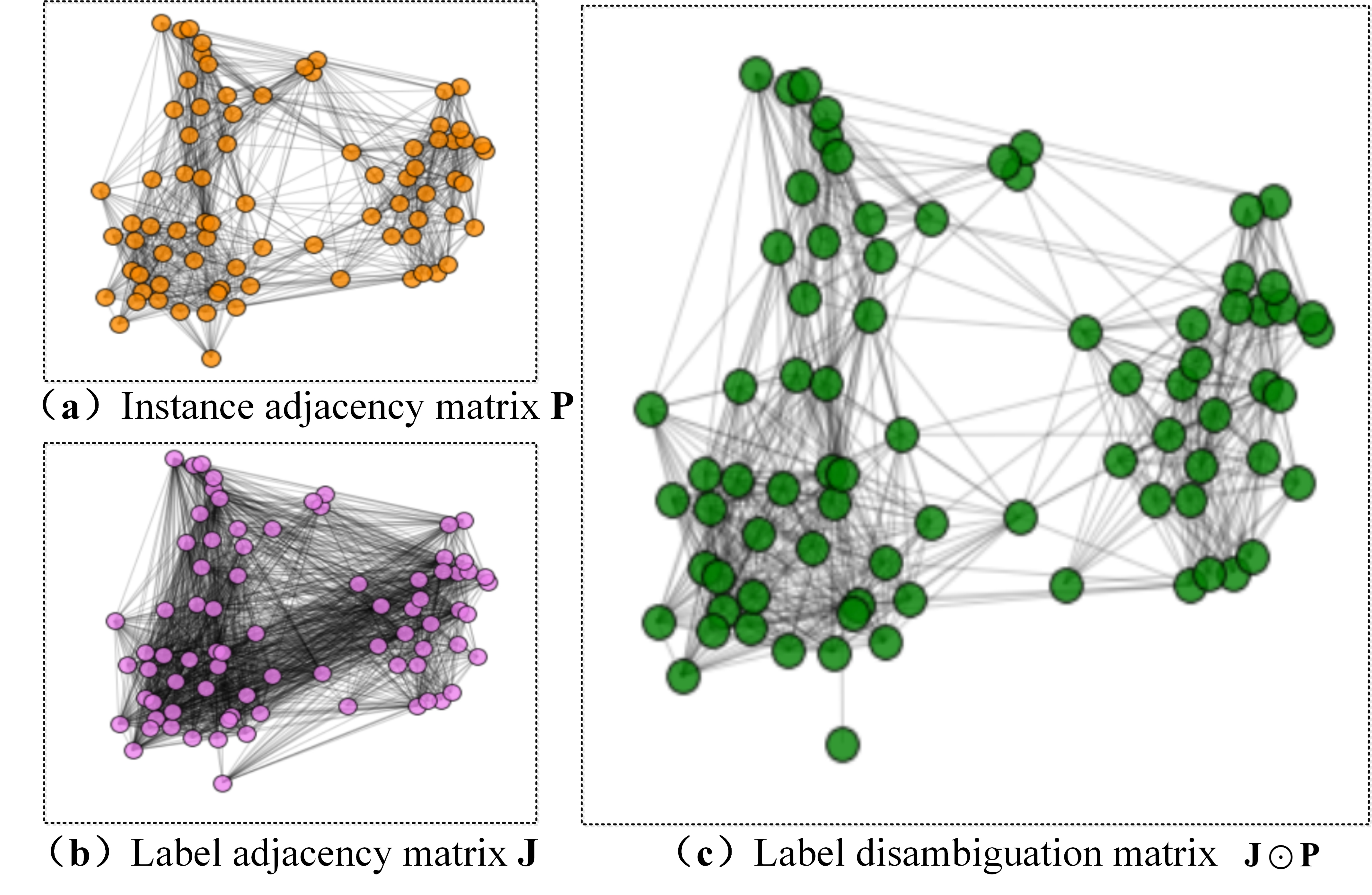}}
		\caption{The visualization of adjacency matrix graph.}
		\label{fig3}
	\end{center}
	\vskip -0.3in
\end{figure}

\subsection{Main Experiment Results} \label{4.4}

\subsubsection{Real-world Partially Labeled Datasets} 
\ 
\indent

\indent The proposed algorithm is evaluated on real-world partially labeled datasets through ten-fold cross-validation to assess its effectiveness. Specifically, \cref{table2} presents the test results of DDMP compared to other algorithms across five real-world datasets.

\begin{itemize}
	\item Across 50 experiments (5 real-world datasets $\times$ 10 comparison algorithms), DDMP significantly outperforms all other methods in $86.0\%$ of the cases.
	\item As shown in \cref{table2}, DDMP achieves superior performance on most real-world datasets compared to other deep learning-based PLL methods. Specifically, on the Lost, MSRCv2, and BirdSong datasets, DDMP outperforms the best competing method in disambiguation accuracy by $0.32\%$, $0.05\%$, $5.66\%$, respectively. This advantage stems from the ability of DDMP to utilize potential information within instances and candidate labels, thereby reducing the effect of ambiguous labels and mitigating the influence of misleading nearest neighbor instances when generating pseudo-clean labels. Additionally, DDMP effectively learns to eliminate the impact of ambiguous labels during the reverse generative process.
	
\end{itemize}

\begin{table*}[t]
	\scriptsize	
	\centering
	\caption{Comparative disambiguation accuracy (mean $\pm$ std) across datasets.}
	\label{tab1}
	\centering
	\begin{tabular}{cccccccc}
		\hline
		\toprule
		\multirow{2}{*}{Datasets} &\multirow{2}{*}{$q$}& \multicolumn{6}{c}{Method} \\  
		\cmidrule(lr){3-8}  
		& & PiCO  & CAVL & VALEN & PRODEN & LW & ABLE \\
		\midrule		
		\multirow{3}{*}{MNIST}  & \textbf{$0.1$} & 98.97 $\pm$ 0.12\% & 98.95 $\pm$ 0.04\%  & 98.75 $\pm$ 0.02\%   & 98.91 $\pm$ 0.03\%  & 98.76 $\pm$ 0.03\%  & 96.72 $\pm$ 0.13\%  \\
	                         	& \textbf{$0.3$} & 98.85 $\pm$ 0.14\% & 98.90 $\pm$ 0.15\%  & 98.66 $\pm$ 0.03\%  & 98.71 $\pm$ 0.02\%   & 98.64 $\pm$ 0.02\%   & 96.50 $\pm$ 0.52\%  \\
	                         	& \textbf{$0.5$} & 98.63 $\pm$ 0.03\% & 98.71 $\pm$ 0.04\%  & 98.32 $\pm$ 0.02\% &98.57 $\pm$ 0.03\%     & 98.41 $\pm$ 0.03\%  & 96.41 $\pm$ 0.06\% \\
		\midrule
		\multirow{3}{*}{Fashion-MNIST}  & \textbf{$0.1$}  & 93.36 $\pm$ 0.09\%   & 90.32 $\pm$ 0.08\%    & 90.72 $\pm$ 0.06\%     & 88.27 $\pm$ 0.03\%   & 88.18 $\pm$ 0.05\% & 92.04 $\pm$ 0.14\%  \\
	                           	& \textbf{$0.3$}        & 93.12 $\pm$ 0.12\%     & 89.77 $\pm$ 0.04\%   & 90.16 $\pm$ 0.06\%    & 87.83 $\pm$ 0.04\%      & 88.02 $\pm$ 0.04\%  & 91.91 $\pm$ 0.09\% \\
	                        	& \textbf{$0.5$}      & 92.88 $\pm$ 0.03\%   & 88.92 $\pm$ 0.11\%  & 89.51 $\pm$ 0.04\%    & 87.01 $\pm$ 0.03\%    & 87.66 $\pm$ 0.06\%   & 91.33 $\pm$ 0.17\% \\
		\midrule
		\multirow{3}{*}{Kuzushiji-MNIST}  & \textbf{$0.1$}   & \textbf{97.68 $\pm$ 0.06}\%    & 93.73 $\pm$ 0.16\%   & 92.67 $\pm$ 0.06\%   & 91.11 $\pm$ 0.03\%    & 92.78 $\pm$ 0.04\%  & 94.89 $\pm$ 0.03\%\\
	                               	& \textbf{$0.3$}         & \textbf{97.34 $\pm$ 0.03}\%    & 93.57 $\pm$ 0.11\%   & 91.75 $\pm$ 0.22\%   & 90.67 $\pm$ 0.04\%   & 91.28 $\pm$ 0.03\% & 94.35 $\pm$ 0.28\% \\
	                            	& \textbf{$0.5$}        & \textbf{97.15 $\pm$ 0.12}\%     & 91.57 $\pm$ 0.22\%    & 90.40 $\pm$ 0.03\%   & 89.00 $\pm$ 0.05\%    & 86.55 $\pm$ 0.06\% & 92.98 $\pm$ 0.31\% \\
		\midrule
		
		\multirow{3}{*}{CIFAR-10}  & \textbf{$0.1$}     & 94.54 $\pm$ 0.05\%     & 83.37 $\pm$ 0.07\%  & 83.21 $\pm$ 0.10\%     & 84.12 $\pm$ 0.36\%   & 81.43 $\pm$ 0.13\% & 92.98 $\pm$ 0.31\% \\
	                        	& \textbf{$0.3$}        & 94.13 $\pm$ 0.08\%   &77.53 $\pm$ 0.34\%  & 82.87 $\pm$ 0.26\%     & 83.36 $\pm$ 0.46\%   & 80.95 $\pm$ 0.17\%   & 92.37 $\pm$ 0.15\%  \\
	                        	& \textbf{$0.5$}        & 93.85 $\pm$ 0.15\%     & 73.25 $\pm$ 0.17\%    & 81.88 $\pm$ 0.20\%   & 77.52 $\pm$ 0.18\%     & 78.72 $\pm$ 0.17\% & 91.67 $\pm$ 0.22\%  \\
		\midrule
		
		\multirow{3}{*}{CIFAR-100}  & \textbf{$0.01$}       & 71.02 $\pm$ 0.29\%   & 45.80 $\pm$ 0.13\%  & 66.79 $\pm$ 0.13\%    & 50.42 $\pm$ 0.28\%  & 51.63 $\pm$ 0.10\%  & 52.81 $\pm$ 0.12\% \\
	                        	& \textbf{$0.05$}        & 70.29 $\pm$ 0.29\%  & 39.87 $\pm$ 0.33\% & 66.15 $\pm$ 0.25\%      & 50.29 $\pm$ 0.29\%     & 46.54 $\pm$ 0.17\%  & 46.26 $\pm$ 0.45\% \\
	                        	& \textbf{$0.1$}         & 58.16 $\pm$ 0.34\%     & 21.55 $\pm$ 0.12\%     & 65.21 $\pm$ 0.33\%  & 46.81 $\pm$ 0.32\%     & 34.56 $\pm$ 0.45\%  & 45.13 $\pm$ 0.21\% \\
		\midrule
		Datasets &$q$ &DIRK  & CRDPLL & CroSel & LS-PLL & DDMP(SimCLR) & DDMP(CLIP) \\
		\midrule

		\multirow{3}{*}{MNIST}  & \textbf{$0.1$}     & 98.92 $\pm$ 0.25\%     & 98.72 $\pm$ 0.22\%  & 98.79 $\pm$ 0.24\%   & 98.83 $\pm$ 0.24\%  & 98.88 $\pm$ 0.21\% & \textbf{99.16 $\pm$ 0.03}\% \\
		& \textbf{$0.3$}                             & 98.54 $\pm$ 0.29\%   & 98.64 $\pm$ 0.19\% & 98.68 $\pm$ 0.31\%     & 98.52 $\pm$ 0.36\%   & 98.76 $\pm$ 0.13\%   & \textbf{99.06 $\pm$ 0.12}\% \\
		& \textbf{$0.5$}                             & 98.31 $\pm$ 0.13\%    & 98.32 $\pm$ 0.33\%    & 98.41 $\pm$ 0.37\%  & 98.39 $\pm$ 0.51\%     & 98.53 $\pm$ 0.09\%   & \textbf{98.90 $\pm$ 0.23}\% \\
		\midrule
		
		\multirow{3}{*}{Fashion-MNIST}  & \textbf{$0.1$}      & 93.71 $\pm$ 0.10\%    & 93.21 $\pm$ 0.24\%  & 93.26 $\pm$ 0.45\%   & 90.49 $\pm$ 0.29\%  & 91.03 $\pm$ 0.32\% & \textbf{93.51 $\pm$ 0.34}\% \\
		& \textbf{$0.3$}                             & 92.10 $\pm$ 0.19\%   & 92.53 $\pm$ 0.31\%   & 93.12 $\pm$ 0.27\%     & 89.78 $\pm$ 0.25\%     & 90.84 $\pm$ 0.14\%  & \textbf{93.24 $\pm$ 0.28}\% \\
		& \textbf{$0.5$}                             & 91.23 $\pm$ 0.24\%    & 91.47 $\pm$ 0.20\%  & 92.87 $\pm$ 0.38\%  &84.09 $\pm$ 0.56\%     & 90.36 $\pm$ 0.41\%   & \textbf{92.97 $\pm$ 0.56}\% \\
		\midrule

		\multirow{3}{*}{Kuzushiji-MNIST}  & \textbf{$0.1$}      & 97.13 $\pm$ 0.32\%     & 97.13 $\pm$ 0.57\%  & 97.07 $\pm$ 0.53\%   & 92.52 $\pm$ 0.37\%  & 97.29 $\pm$ 0.13\%  & 94.01 $\pm$ 0.17\%  \\
		& \textbf{$0.3$}                             & 96.49 $\pm$ 0.47\%     & 96.55 $\pm$ 0.45\%  & 96.44 $\pm$ 0.38\%     & 86.38 $\pm$ 0.34\%    & 96.86 $\pm$ 0.08\%   & 93.89 $\pm$ 0.24\% \\
		& \textbf{$0.5$}                             & 96.13 $\pm$ 0.51\%     & 96.21 $\pm$ 0.61\%  & 95.96 $\pm$ 0.27\%  & 78.85 $\pm$ 0.54\%     & 96.35 $\pm$ 0.21\%  & 93.45 $\pm$ 0.31\% \\
		\midrule

		\multirow{3}{*}{CIFAR-10}  & \textbf{$0.1$}     & 92.41 $\pm$ 0.23\%    & 94.71 $\pm$ 0.18\%   & 97.29 $\pm$ 0.13\%     & 84.31 $\pm$ 0.45\%   & 93.02 $\pm$ 0.32\%  & \textbf{97.79 $\pm$ 0.12}\% \\
		& \textbf{$0.3$}        & 92.14 $\pm$ 0.12\% & 94.27 $\pm$ 0.24\%   & 92.47 $\pm$ 0.61\%     & 79.05 $\pm$ 0.52\%  & 92.57 $\pm$ 0.41\%  & \textbf{97.71 $\pm$ 0.17}\% \\
		& \textbf{$0.5$}         & 91.34 $\pm$ 0.14\%   & 93.84 $\pm$ 0.11\%     & 91.97 $\pm$ 0.45\%    & 67.82 $\pm$ 0.36\%       & 91.89 $\pm$ 0.27\% & \textbf{97.36 $\pm$ 0.23}\%  \\
		\midrule
		
		\multirow{3}{*}{CIFAR-100}  & \textbf{$0.01$}     & 67.71 $\pm$ 0.51\%  & 76.70 $\pm$ 0.23\%  & 76.70 $\pm$ 0.27\%    & 53.45 $\pm$ 0.34\%     & 69.11 $\pm$ 0.43\%  & \textbf{84.03 $\pm$ 0.05}\% \\
		& \textbf{$0.05$}        & 64.47 $\pm$ 0.32\%     & 76.67 $\pm$ 0.18\%    & 75.71 $\pm$ 0.46\%    & 46.23 $\pm$ 0.45\%  & 68.13 $\pm$ 0.23\% & \textbf{83.37 $\pm$ 0.16}\% \\
		& \textbf{$0.1$}        & 64.28 $\pm$ 0.47\%    & 71.21 $\pm$ 0.34\%      & 74.92 $\pm$ 0.37\%    & 32.78 $\pm$ 0.74\%     & 66.29 $\pm$ 0.52\%  & \textbf{81.65 $\pm$ 0.09}\% \\
				
		\bottomrule
		\hline		
	\end{tabular}
\end{table*}

\begin{table*}[t]
    \scriptsize
    \centering
    \caption{Comparison of different PLL methods on instance-dependent PLL datasets (mean $\pm$ std).}
    \label{tab:instance-analysis}
    \resizebox{\textwidth}{!}{
    \begin{tabular}{lccccccccc}
        \toprule
        \multirow{2}{*}{Datasets} & \multicolumn{9}{c}{Method} \\
        \cmidrule(lr){2-10}
        & PiCO & CAVL & VALEN & PRODEN & LW & ABLE & DIRK & DDMP (SimCLR) & DDMP (CLIP) \\
        \midrule

        MNIST
        & 98.61 $\pm$ 0.12\% & 98.84 $\pm$ 0.05\% & 98.72 $\pm$ 0.05\% & 98.39 $\pm$ 0.10\% & 98.56 $\pm$ 0.06\% & 98.82 $\pm$ 0.12\% & 99.04 $\pm$ 0.07\% & 98.89 $\pm$ 0.09\% & \textbf{99.10 $\pm$ 0.07\%} \\
        
        Fashion-MNIST
        & 88.41 $\pm$ 0.20\% & 87.94 $\pm$ 0.19\% & 90.63 $\pm$ 0.30\% & 89.79 $\pm$ 0.24\% & 88.99 $\pm$ 0.26\% & 91.43 $\pm$ 0.08\% & 91.48 $\pm$ 0.21\% & 90.06 $\pm$ 0.15\% & \textbf{92.82 $\pm$ 0.17\%} \\
        
        Kuzushiji-MNIST
        & 94.78 $\pm$ 0.19\% & 93.69 $\pm$ 0.28\% & 96.19 $\pm$ 0.75\% & 93.79 $\pm$ 0.24\% & 92.27 $\pm$ 1.03\% & \textbf{98.17 $\pm$ 0.02\%} & 96.80 $\pm$ 0.52\% & 96.19 $\pm$ 0.17\% & 90.79 $\pm$ 0.21\% \\
        
        CIFAR-10
        & 86.18 $\pm$ 0.21\% & 59.67 $\pm$ 3.30\% & 85.48 $\pm$ 0.62\% & 86.04 $\pm$ 0.21\% & 37.49 $\pm$ 2.82\% & 87.62 $\pm$ 0.67\% & 90.87 $\pm$ 0.25\% & 90.66 $\pm$ 0.19\% & \textbf{96.51 $\pm$ 0.10\%} \\
        
        CIFAR-100
        & 62.98 $\pm$ 0.38\% & 52.59 $\pm$ 1.01\% & 62.96 $\pm$ 0.96\% & 62.56 $\pm$ 1.49\% & 53.98 $\pm$ 0.99\% & 65.43 $\pm$ 0.39\% & 68.77 $\pm$ 0.49\% & 67.29 $\pm$ 0.27\% & \textbf{81.61 $\pm$ 0.33\%} \\
        
        \bottomrule
    \end{tabular}
    }
\end{table*}

\begin{table*}[t]
    \scriptsize
    \centering
    \caption{Comparison of different encoders and KNN-only baselines (mean $\pm$ std) across datasets.}
    \label{tab:encoder-analysis}
    \begin{tabular}{ccccccc}
        \hline
        \toprule
        \multirow{2}{*}{Datasets} & \multirow{2}{*}{$q$} & \multicolumn{5}{c}{Method} \\
        \cmidrule(lr){3-7}
        & & DDMP (ResNet) & SimCLR KNN & DDMP (SimCLR) & CLIP KNN & DDMP (CLIP) \\
        \midrule

        \multirow{3}{*}{MNIST}
        & \textbf{0.1} & 98.21 $\pm$ 0.05\% & 97.42 $\pm$ 0.07\% & 98.88 $\pm$ 0.21\% & 97.71 $\pm$ 0.23\% & \textbf{99.16 $\pm$ 0.03\%} \\
        & \textbf{0.3} & 98.09 $\pm$ 0.10\% & 97.27 $\pm$ 0.05\% & 98.76 $\pm$ 0.13\% & 97.36 $\pm$ 0.14\% & \textbf{99.06 $\pm$ 0.12\%} \\
        & \textbf{0.5} & 97.93 $\pm$ 0.07\% & 96.05 $\pm$ 0.16\% & 98.53 $\pm$ 0.09\% & 96.38 $\pm$ 0.09\% & \textbf{98.90 $\pm$ 0.23\%} \\
        \midrule

        \multirow{3}{*}{Fashion-MNIST}
        & \textbf{0.1} & 89.45 $\pm$ 0.09\% & 86.20 $\pm$ 0.13\% & 91.03 $\pm$ 0.32\% & 87.98 $\pm$ 0.13\% & \textbf{93.51 $\pm$ 0.34\%} \\
        & \textbf{0.3} & 88.73 $\pm$ 0.14\% & 84.98 $\pm$ 0.20\% & 90.84 $\pm$ 0.14\% & 86.96 $\pm$ 0.23\% & \textbf{93.24 $\pm$ 0.28\%} \\
        & \textbf{0.5} & 88.02 $\pm$ 0.19\% & 82.26 $\pm$ 0.09\% & 90.36 $\pm$ 0.41\% & 83.84 $\pm$ 0.17\% & \textbf{92.97 $\pm$ 0.56\%} \\
        \midrule

        \multirow{3}{*}{Kuzushiji-MNIST}
        & \textbf{0.1} & 95.73 $\pm$ 0.13\% & 69.44 $\pm$ 0.09\% & \textbf{97.29 $\pm$ 0.13\%} & 78.67 $\pm$ 0.17\% & 94.01 $\pm$ 0.17\% \\
        & \textbf{0.3} & 95.17 $\pm$ 0.09\% & 66.81 $\pm$ 0.15\% & \textbf{96.86 $\pm$ 0.08\%} & 75.45 $\pm$ 0.31\% & 93.89 $\pm$ 0.24\% \\
        & \textbf{0.5} & 94.29 $\pm$ 0.21\% & 62.08 $\pm$ 0.25\% & \textbf{96.35 $\pm$ 0.21\%} & 69.83 $\pm$ 0.21\% & 93.45 $\pm$ 0.31\% \\
        \midrule

        \multirow{3}{*}{CIFAR-10}
        & \textbf{0.1} & 89.28 $\pm$ 0.23\% & 88.61 $\pm$ 0.12\% & 93.02 $\pm$ 0.32\% & 91.87 $\pm$ 0.06\% & \textbf{97.79 $\pm$ 0.12\%} \\
        & \textbf{0.3} & 88.91 $\pm$ 0.15\% & 86.59 $\pm$ 0.23\% & 92.57 $\pm$ 0.41\% & 90.20 $\pm$ 0.29\% & \textbf{97.71 $\pm$ 0.17\%} \\
        & \textbf{0.5} & 88.05 $\pm$ 0.31\% & 81.77 $\pm$ 0.05\% & 91.89 $\pm$ 0.27\% & 86.91 $\pm$ 0.34\% & \textbf{97.36 $\pm$ 0.23\%} \\
        
        \hline
    \end{tabular}
\end{table*}

\begin{figure}[t]
	\vskip 0in
	\begin{center}
		\centerline{\includegraphics[width=0.68\linewidth]{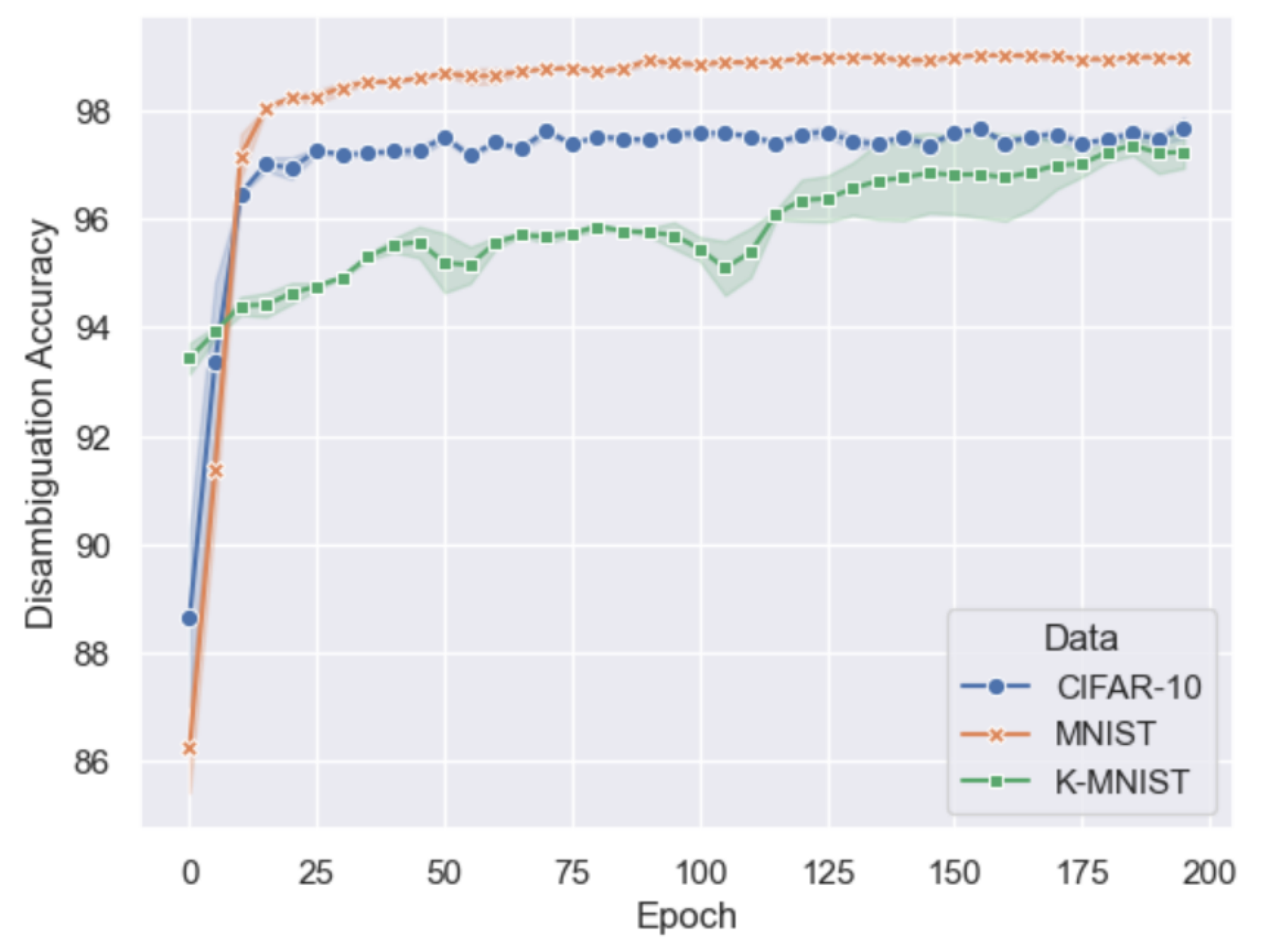}}
		\caption{Convergence results of DDMP on benchmark datasets.}
		\label{fig4}
	\end{center}
	\vskip -0.3in
\end{figure}

\begin{figure*}[t] 
	\centering
	\includegraphics[width=0.95\linewidth]{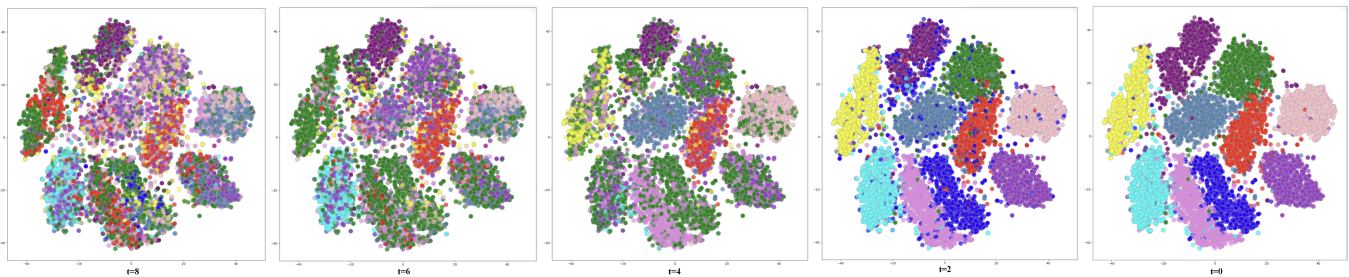}
	\vskip -0.1in
	\caption{T-SNE visualization of MNIST dataset during DDMP reverse generation process.}
	\label{fig5}
	\vskip -0.1in
\end{figure*}

\subsubsection{Synthesized Partially Labeled Datasets} 
\ 
\indent

The comparison algorithms are evaluated on various synthesized benchmark datasets to thoroughly validate the effectiveness of DDMP. Despite comparison algorithms exhibiting specialized strengths per dataset in \cref{tab1}, DDMP achieves universal performance dominance, outperforming all comparative methods in most cases.

\begin{itemize}
	\item 
	As illustrated in \cref{tab1}, DDMP consistently achieves better disambiguation results across datasets with different degrees of label ambiguity. Specifically, out of 150 experiments (10 comparison algorithms $\times$ 15 synthesized benchmark datasets), DDMP(CLIP) significantly outperforms the comparative methods in $91.3\%$ cases. Meanwhile, DDMP(SimCLR) is $70.0\%$ higher than comparative methods and achieves comparable performance to PiCO.
	\item 
	‌Across all evaluated synthesized benchmark datasets, DDMP(CLIP) demonstrates consistently superior performance relative to other deep learning PLL methods. Notably, on the CIFAR-10 and CIFAR-100 datasets,  it is significantly better than the best comparison algorithm. 
	\item 	
	When compared with deep learning-based PLL methods, DDMP(SimCLR) consistently performs better than VALEN, PRODEN, and LW in most instances. Specifically, on the Kuzushiji-MNIST, CIFAR-10 and CIFAR-100 datasets, DDMP(SimCLR) outperforms the best deep learning PLL method by $3.56\%$, $9.65\%$, $2.32\%$.	
	\item 
	It is important to highlight that DDMP achieves high performance on synthetic benchmarks and remains robust even as the value of $q$ grows, further validating its effectiveness in addressing the PLL problem. Moreover, for the CLIP pre-training model, this paper only used the instance encoder, and did not use the label encoder. For DDMP, by comparing the two pre-training models of SimCLR and CLIP, it can be found that good encoding features can significantly improve the disambiguation performance of ambiguous labels in PLL. This insight offers a promising direction for future research, suggesting that the strengths of unsupervised contrastive learning can be harnessed to mitigate the impact of ambiguous labels.
\end{itemize}

\subsubsection{Instance-dependent Partially Labeled Datasets}
\indent

To further evaluate the robustness of DDMP under instance-dependent candidate-label ambiguity, we additionally compare DDMP with both conventional PLL baselines and representative instance-dependent PLL methods on instance-dependent PLL datasets. Following the commonly used construction strategy in VALEN, these datasets are generated such that the candidate-label ambiguity depends on the instance rather than only on the class label, thus providing a more challenging and realistic evaluation setting.

\begin{itemize}
    \item As shown in \cref{tab:instance-analysis}, DDMP still achieves competitive performance on instance-dependent PLL datasets, even though it is not specifically designed for this setting. This result suggests that the effectiveness of DDMP is not restricted to the standard instance-independent synthetic protocol, and that the proposed framework remains robust when the ambiguity mechanism becomes more complex.
    
    \item Across the five instance-dependent PLL datasets, DDMP equipped with CLIP features achieves the best performance on four datasets, namely MNIST, Fashion-MNIST, CIFAR-10, and CIFAR-100, with accuracies of $99.10\%$, $92.82\%$, $96.51\%$, and $81.61\%$, respectively. In particular, the performance gains on CIFAR-10 and CIFAR-100 are especially notable, showing that DDMP can still effectively reduce ambiguity under more challenging instance-dependent supervision when stronger semantic features are available.
    
    \item DDMP with SimCLR features also demonstrates strong effectiveness on instance-dependent PLL datasets. Although DDMP(CLIP) achieves the best overall performance on most datasets, DDMP(SimCLR) remains competitive and even outperforms DDMP(CLIP) on Kuzushiji-MNIST. This observation indicates that the effectiveness of DDMP does not stem solely from the use of a stronger encoder such as CLIP. Instead, the proposed diffusion-based progressive disambiguation framework itself plays the central role, while different pre-trained encoders may exhibit different strengths across datasets.
    
    \item Compared with representative instance-dependent PLL methods such as VALEN, ABLE, and DIRK, DDMP shows favorable robustness on several datasets, especially MNIST, Fashion-MNIST, CIFAR-10, and CIFAR-100. At the same time, the results on Kuzushiji-MNIST indicate that methods explicitly designed for instance-dependent ambiguity may still have advantages in certain cases. Overall, these results support our claim that the assumption used in the generative formulation of DDMP mainly serves as a modeling approximation rather than a strict requirement for practical effectiveness.
\end{itemize}

\subsubsection{ Effect of Pre-trained Encoders and KNN-only Baselines} 
\indent

\indent To further examine whether the performance gains of DDMP mainly come from strong pre-trained encoders, we additionally compare DDMP under different encoders and report KNN-only baselines that remove the diffusion disambiguation framework.
\begin{itemize}
	\item As shown in \cref{tab:encoder-analysis}, stronger pre-trained encoders such as CLIP do improve the upper bound of PLL performance. However, the gains of DDMP cannot be attributed solely to the encoder itself. When the pre-trained encoder is replaced by a standard ResNet backbone, DDMP still achieves strong performance across multiple datasets. For example, at $q=0.3$, DDMP (ResNet) reaches 98.09\% on MNIST, 88.73\% on Fashion-MNIST, 95.17\% on Kuzushiji-MNIST, and 88.91\% on CIFAR-10, showing that DDMP remains effective even without relying on a powerful pre-trained encoder.
	\item More importantly, using a strong encoder alone is insufficient to achieve the performance of DDMP. The SimCLR KNN and CLIP KNN baselines, which only use pre-trained representations together with KNN but remove the diffusion disambiguation framework, perform consistently worse than DDMP with the same encoder. This gap is especially clear on difficult datasets such as Kuzushiji-MNIST and CIFAR-10, and it further enlarges as the noise rate increases. Therefore, the superior performance of DDMP does not simply come from feature extraction quality, but from the proposed diffusion-based progressive disambiguation mechanism.
    \item This observation is also consistent with the real-world partially labeled experiments, where the processed data are directly used as encoding features rather than relying on a pre-trained encoder, while DDMP still outperforms competing methods in most cases. Together with the ablation results on the information complementarity strategy and the transition-aware matrix, these results further demonstrate that the performance gains of DDMP mainly come from its task-specific ambiguity reduction framework rather than the backbone alone.
\end{itemize}

To better verify the disambiguation effectiveness of DDMP, we tested the performance of different PLL methods on a large number of datasets. Despite the growing label ambiguity with increasing $q$, DDMP consistently delivers superior disambiguation performance, surpassing nearly all other compared PLL methods. This impressive result can be attributed to the integration of the information complementarity label disambiguation strategy and the estimation of the transition-aware matrix, which together effectively extract valuable information from instances and candidate labels, enabling continuous refinement of the label matrix and mitigating the influence of ambiguity. Specifically, DDMP reformulates the label disambiguation problem from the perspective of generative models, where labels are generated by iteratively refining initial random guesses. This perspective enables the diffusion model to learn how label information is generated stochastically, effectively mitigating the impact of ambiguous labeling. It is also noteworthy that pre-trained instance encoders remain unaffected by label ambiguity, as they are trained via self-supervised learning and significantly enhance the adversarial robustness of the model\cite{hendrycks2019using}\cite{chen2024label}.

\begin{figure*}[ht] 
	\centering
	\subfigure[CIFAR-10 dataset, $q=0.3$]{
		\label{fig:subfig:a} 
		\includegraphics[width=0.83\linewidth]{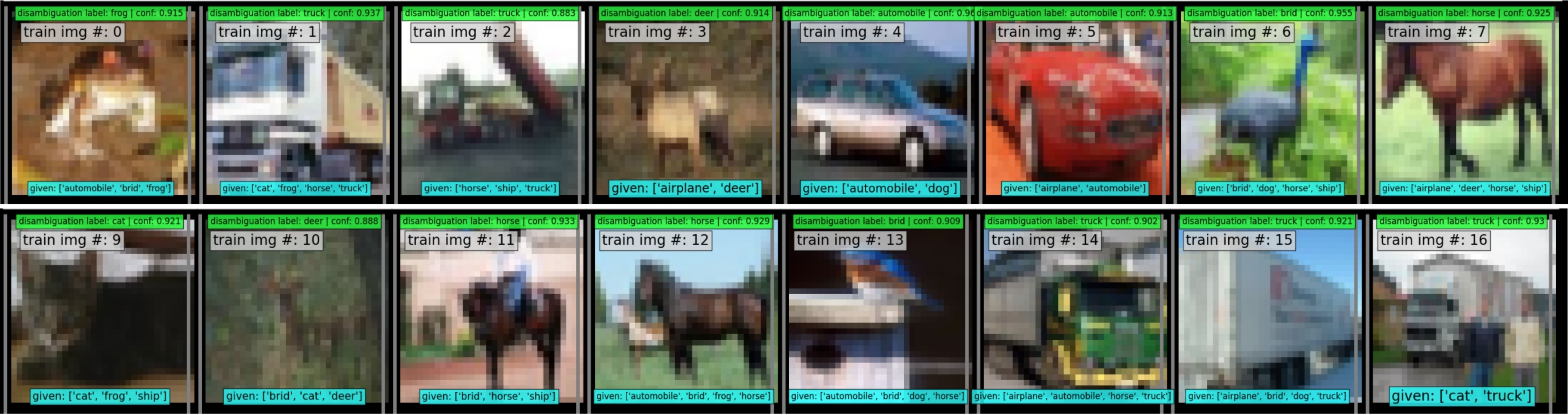}}
	
	\subfigure[Kuzushiji-MNIST dataset, $q=0.3$]{
		\label{fig:subfig:b} 
		\includegraphics[width=0.83\linewidth]{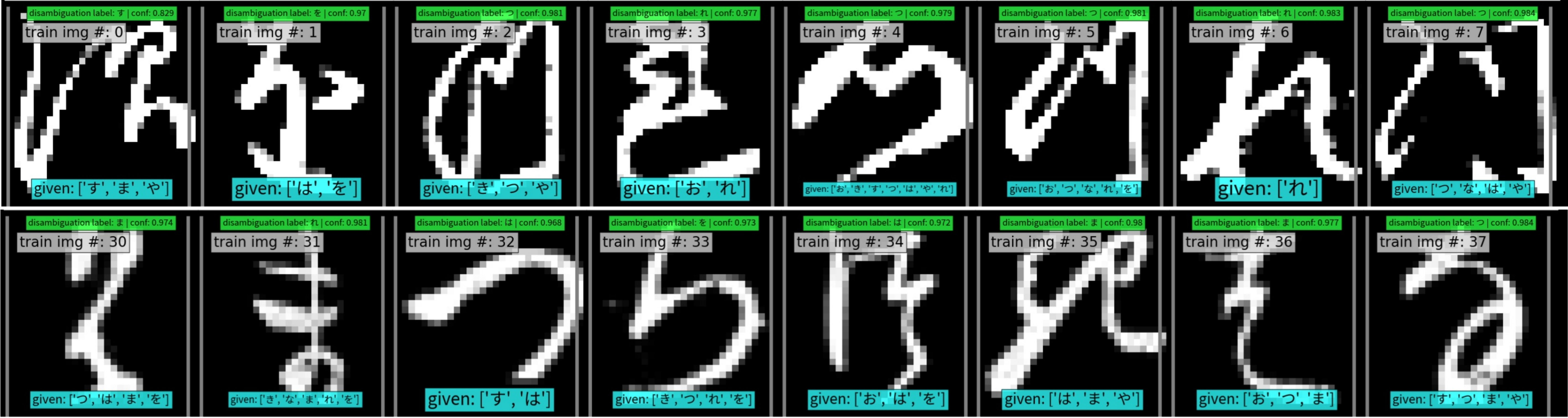}}
	\caption{Visual Representation of Disambiguation Results from the DDMP Algorithm.}
	\label{fig6} 
\end{figure*}

\subsection{Ablation Study}

To better assess the role of each component within the DDMP method, ablation experiments were performed on two specific parts, i.e., the information complementarity label disambiguation strategy and the estimation of the transition-aware matrix  are also conducted. From the ablation results presented in \cref{tab3}, the following conclusions can be derived.




\begin{itemize}
	\item Compared with DDMP-w/o-$\mathbf{I}$, DDMP shows a good disambiguation ability. This is attributed to the complementary information between instances and candidate labels, which helps mitigate the influence of ambiguous labels and effectively guides learning process of the model. To better illustrate this point, the adjacency matrices are visualized. From \cref{fig3}, it can be seen that the connection lines in points $\mathbf{P}$ and $\mathbf{J}$ are significantly reduced, as the potential information of candidate labels can effectively reduce the impact of negative nearest neighbor instance in the disambiguation process, thereby learning accurate pseudo-clean labels.
	\item DDMP achieves better disambiguation results compared to DDMP-w/o-$\mathbf{T}$, as the label transition-aware matrix helps mitigate the effects of ambiguous labels by capturing the underlying distribution of the true labels.
	\item The variant DDMP-w/o-$\mathbf{IT}$ suffers a slight drop in performance. This is attributed to the mutual reinforcement between the initial label disambiguation and the transition-aware matrix, which together enhance the ability of the model to resolve label ambiguity. As training advances, the model progressively uncovers the true labels, and these increasingly accurate labels further steer the learning process, leading to improved disambiguation performance.
\end{itemize}

\begin{table}
	\centering
	\caption{Ablation study of DDMP(SimCLR) on CIFAR-10 ($q=0.3$) and BirdSong datasets.}
	\label{tab3}
	\begin{tabular}{c|cc|cc} 
		\hline
		\toprule
		\textbf{Ablation} & \textbf{$\mathbf{I}$} & \textbf{$\mathbf{T}$} & CIFAR-10  & BirdSong \\
		\midrule
		DDMP & $\usym{1F5F8}$ & $\usym{1F5F8}$ & 92.57 $\pm$ 0.41\% & 78.96 $\pm$ 0.42\%\\  
		DDMP-w/o-$\mathbf{I}$ & $\usym{2717}$ & $\usym{1F5F8}$ & 88.34 $\pm$ 0.07\% & 69.05 $\pm$ 0.64\%\\  
		DDMP-w/o-$\mathbf{T}$ & $\usym{1F5F8}$ & $\usym{2717}$ & 88.79 $\pm$ 0.34\% & 72.43 $\pm$ 0.38\%\\ 
		DDMP-w/o-$\mathbf{IT}$ & $\usym{2717}$ & $\usym{2717}$ & 78.77 $\pm$ 0.32\% & 67.17 $\pm$ 0.36\%\\ 
		\bottomrule 
		\hline
	\end{tabular}
\end{table}

\begin{figure}[t]
    \centering
    \subfigure[Synthesized PLL datasets]{
        \includegraphics[width=0.45\linewidth]{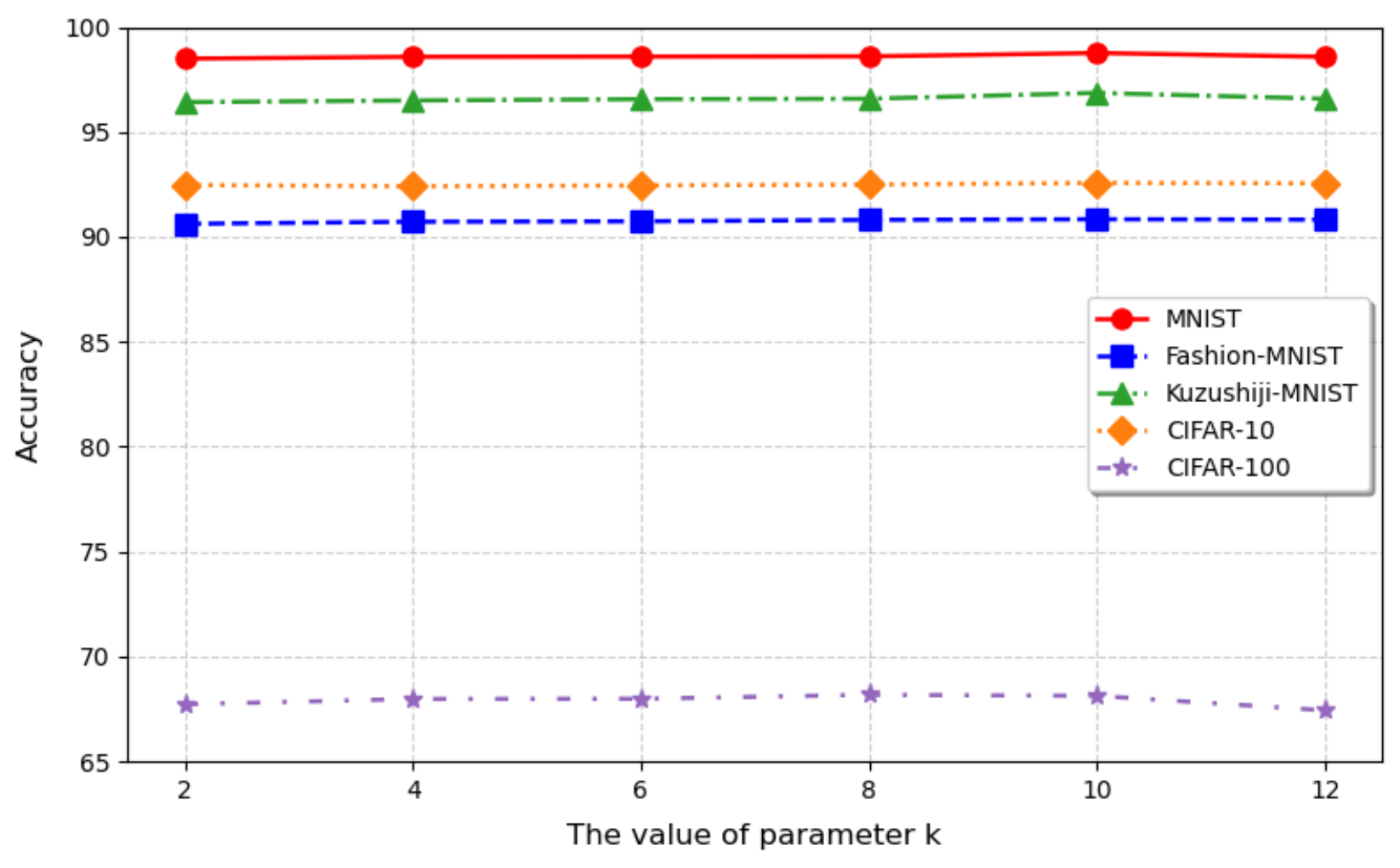}
        \label{fig7a}
    }
    \hfill
    \subfigure[Real-World PLL datasets]{
        \includegraphics[width=0.45\linewidth]{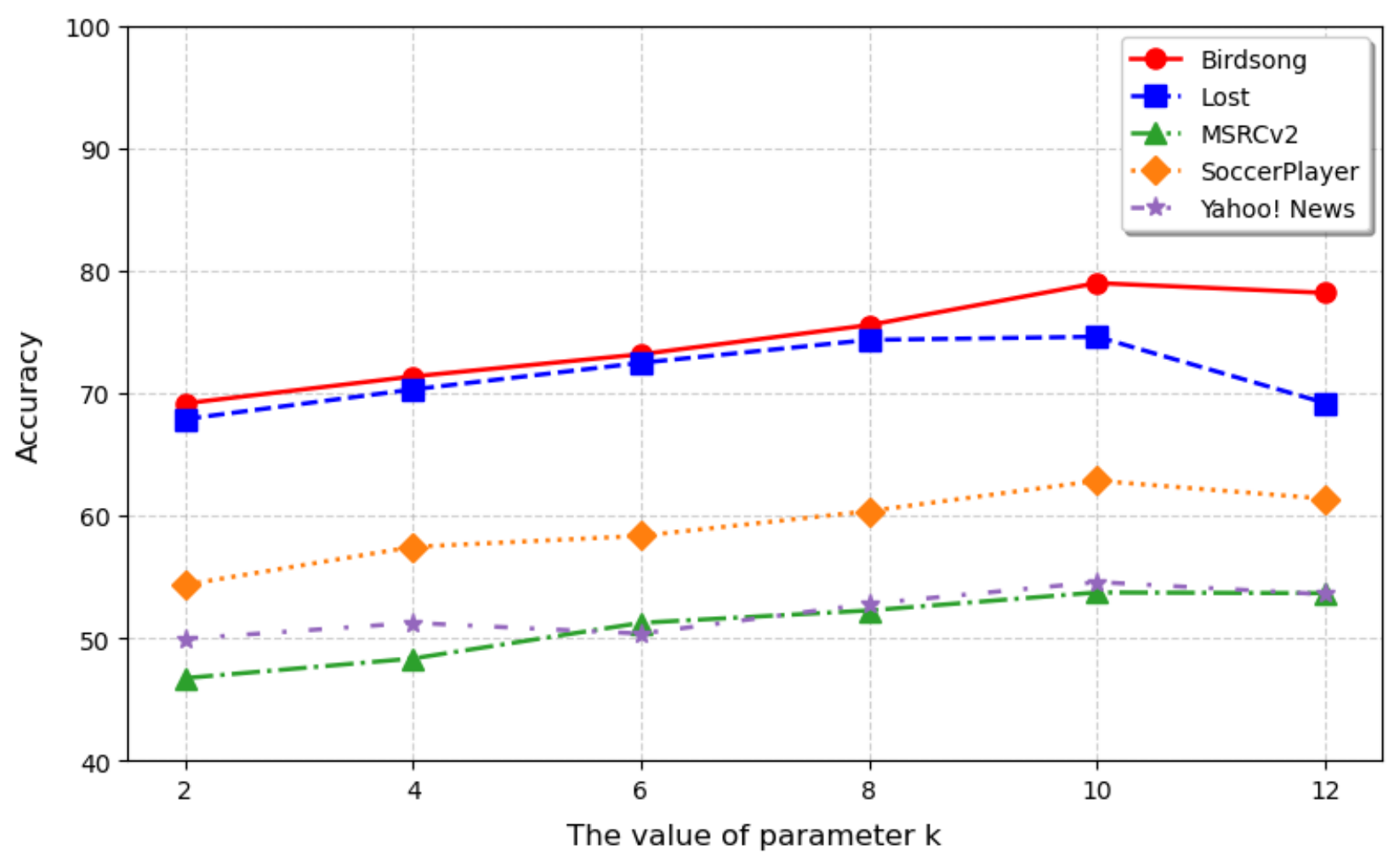}
        \label{fig7b}
    }
    \caption{Parameter sensitivity analysis of $k$ on Synthesized and Real-World PLL datasets.}
    \label{fig7}
\end{figure}

\subsection{Convergence Analysis}


To more intuitively demonstrate the convergence of DDMP, we analyze its performance on three synthetic benchmark datasets, specifically using datasets with $q = 0.3$. As shown in \cref{fig4}, DDMP achieves effective label disambiguation on these benchmarks. As training proceeds, the model steadily approaches a stable performance level, which empirically verifies its convergence. Additionally, the results on both synthetic and real-world datasets indicate that DDMP not only excels in disambiguation but also maintains strong stability throughout the training process.

\subsection{Visualization Analysis} \label{4.5}

To more intuitively reflect the superiority of DDMP, \cref{fig5} demonstrates the visualization of the reverse denoising and disambiguation process of DDMP, which transforms the latent distribution into the ground-truth distribution of labels through a series of multi-step inverse operations. As shown in \cref{fig5}, each cluster achieved accurate classification, which also verifies the effectiveness of the DDMP algorithm in label denoising and disambiguation. To further highlight the advantages of the DDMP algorithm, disambiguation visualization is conducted on the CIFAR-10 and Kuzushiji-MNIST datasets with a noise level of $q = 0.3$. As illustrated in \cref{fig6}, the blue boxes indicate the candidate label sets for each instance, while the green boxes represent the disambiguation results along with the confidence scores for the corresponding label classes. Therefore, it is evident that the DDMP can effectively achieve accurate disambiguation even in the presence of a significant number of ambiguous labels. 


\subsection{Computational Complexity} 

\begin{table}[t]
\centering
\caption{Computational cost comparison with representative PLL methods.}
\label{tab:time-complexity}
\resizebox{\columnwidth}{!}{
\begin{tabular}{lccc}
\toprule
Method & Training FLOPs & Inference FLOPs & Training Time (min/epoch) \\
\midrule
DDMP   & 13.650 GFLOPs & 3.162 GFLOPs & 0.65 \\
PiCO   & 10.289 GFLOPs & 1.114 GFLOPs & 0.60 \\
LS-PLL & 3.813 GFLOPs  & 1.117 GFLOPs & 0.54 \\
CAVL   & 0.567 GFLOPs  & 0.141 GFLOPs & 0.48 \\
\bottomrule
\end{tabular}}
\end{table}

 For the information complementarity label disambiguation strategy, the time complexity of the pseudo-clean label matrix can be expressed as $\mathcal{O}(N^{2}d+N^{2}Q)$. The time complexity of DDMP for one diffusion disambiguation learning is $\mathcal{O}(N\cdot C_{\theta})$, where $C_{\theta}$ represents the cost of calculating the unit sample through the DDMP predictor in both forward and backward directions. Furthermore, the time complexity of update  transition-aware matrix $\mathbf{T}$ and pseudo-clean label matrix $\mathbf{S}$ is $\mathcal{O}(NQ^{2}+Q^{3})$. Therefore, the complexity of DDMP single round training is $\mathcal{O}(N\cdot C_{\theta}+NQ^{2}+Q^{3})$. During the training phase, DDMP only samples a single $t$ in each iteration and does not grow linearly according to $T$. In the inference stage, it is necessary to execute $T$ times along the diffusion chain, with a time complexity of $\mathcal{O}(10 \cdot T \cdot C_{\theta}^{fwd})$, where $C_{\theta}^{fwd}$ represents the cost of one forward propagation. 

To further quantify the practical computational overhead of DDMP, we compare it with other PLL methods in terms of training FLOPs, inference FLOPs, and training time per epoch under the same experimental environment on an NVIDIA RTX 4090 GPU. As shown in \cref{tab:time-complexity}, DDMP requires 13.650 GFLOPs for training and 3.162 GFLOPs for inference, with a training time of 0.65 min/epoch on CIFAR-10 with $q=0.3$. Admittedly, DDMP incurs higher computational cost than several discriminative PLL methods, mainly due to the iterative denoising procedure introduced by diffusion modeling. However, this overhead is effectively controlled through practical design choices, including DDIM-based accelerated sampling and a lightweight noise-prediction network. As a result, the runtime of DDMP remains close to that of PiCO (0.65 vs.\ 0.60 min/epoch), while achieving better empirical performance. These results indicate that DDMP maintains its computational cost within a practically acceptable range and achieves a favorable trade-off between efficiency and disambiguation performance.

\subsection{Parameter Sensitivity Analysis} 

The DDMP involves main parameters, i.e., the nearest neighbors coefficient $k$ and the number of epochs. As shown in \cref{fig4}, the performance of DDMP first gets better and better as the number of iterations increases, and then does not change anymore when it approaches a certain value. Therefore, we set the epoch to 200 in the DDMP algorithm. \cref{fig7} shows the sensitivity of DDMP to the nearest-neighbor coefficient $k$ on both synthesized and real-world PLL datasets. On the synthesized datasets (\cref{fig7a}), DDMP obtains the best overall performance at $k=10$, while the variation among nearby choices remains relatively small, demonstrating the robustness of DDMP to moderate changes in $k$. On the real-world datasets (\cref{fig7b}), the superiority of $k=10$ becomes more evident. This observation suggests that $k=10$ yields a good trade-off between preserving useful local neighborhood structure and avoiding the introduction of irrelevant or misleading neighbors. These results further support the effectiveness of leveraging complementary information from instances and candidate labels to reduce ambiguity and facilitate model learning.

\subsection{Uncertainty Quantification of PLL Methods} \label{4.6}

\begin{figure*}[t] 
	\centering
	\includegraphics[width=0.70\linewidth]{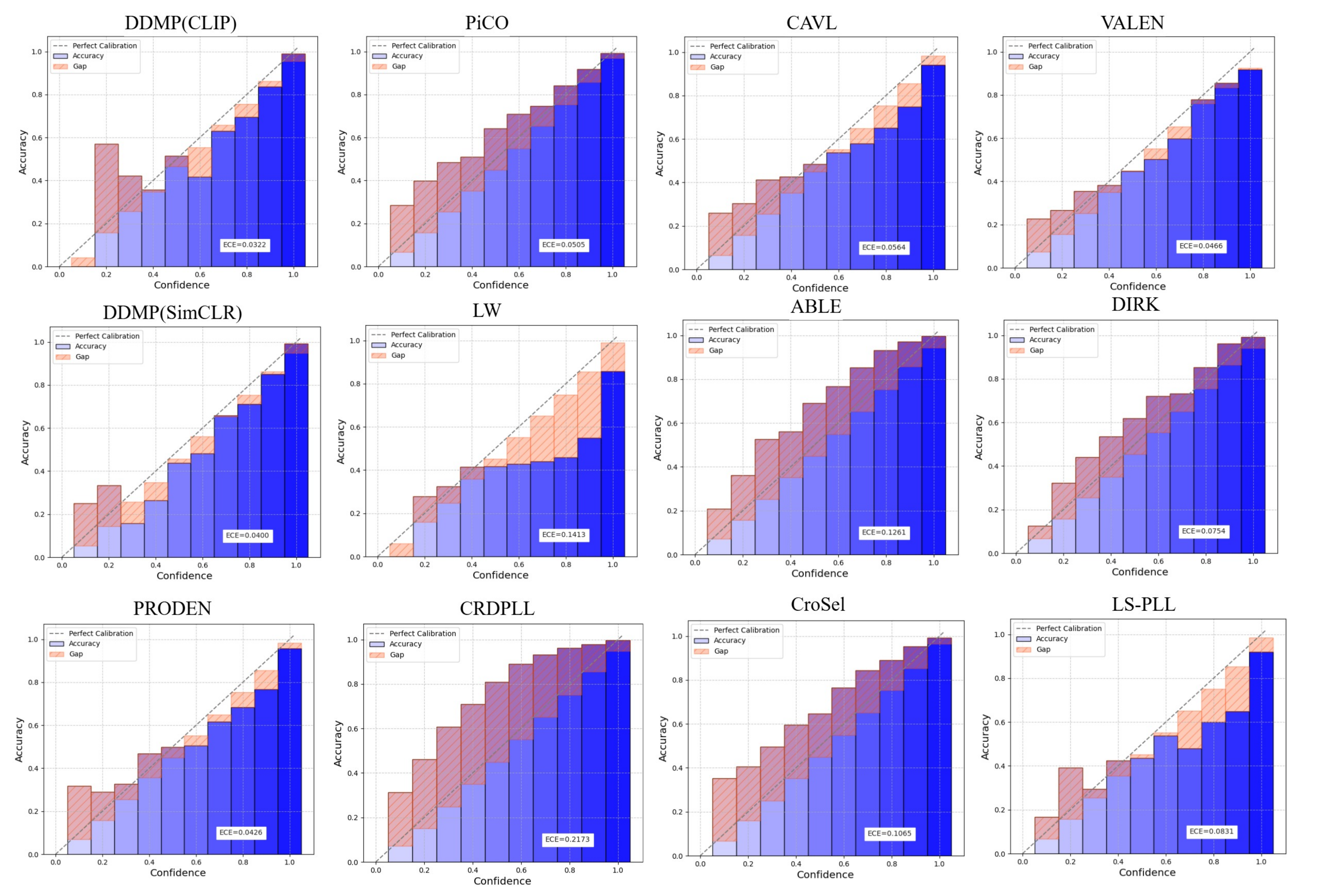}
	\vskip -0.1in
	\caption{Reliability diagram and expected calibration error.}
	\label{fig8}
	\vskip -0.1in
\end{figure*}

When deploying machine learning systems, it is essential that algorithms are not only accurate but also reliable and capable of recognizing potential errors, as this better reflects the effectiveness and reliability of the model  \cite{guo2017calibration}\cite{wu2024distilling}. Therefore, we conducted a preliminary analysis of the output uncertainty associated with various PLL methods, as illustrated in \cref{fig8}. The results indicate that DDMP enables the diffusion model to learn how label information is randomly generated by mining potential valid information between instances. This not only effectively reduces the impact of label ambiguity, but also makes the model more robust and reliable, thereby improving the credibility of the model output and achieving competitive \emph{expected calibration error} (ECE) scores. Specifically, our method achieves the lowest ECE score among the comparison methods on CIFAR-10. The analysis reveals a near-perfect alignment between actual accuracy and average consistency, both of which closely follow the ideal calibration line. The extremely low expected calibration error demonstrates excellent calibration performance. Especially within the high consistency range (0.6-1.0), the deviation between consistency and accuracy can be ignored, indicating an extremely accurate reliability assessment of high certainty predictions. Although the sample size is small in the low consistency range (0.0-0.4), the calibration quality is still strong with only slight statistical fluctuations. A significant feature appears in the data distribution, the actual accuracy strictly monotonically improves over the consistency interval, and the deviation of the accuracy from the predicted consistency in each interval is minimal. This strong positive correlation indicates superior probabilistic calibration. Most notably, in the high confidence interval of (0.8-1.0), the actual accuracy is close to the theoretical maximum, which provides convincing evidence for the reliability of the model in high-certainty predictions.

\subsection{Limitation and Future Work}

Although extensive experiments have demonstrated the effectiveness of DDMP, there are some limitations in practical application. The initial k-NN graph is kept fixed during training, DDMP progressively refines the pseudo-clean label distributions through transition-aware updating and reverse denoising. Experimental results on both real-world and synthesized partially labeled datasets demonstrate the practical effectiveness of this refinement mechanism. Although DDMP has some effectiveness in mitigating noise labeling in negative nearest neighbor instances, it can still lead to the continuous accumulation of erroneous pseudo labels when the initial neighbor relationship is severely affected by noise (e.g., due to a low-quality feature space). In addition, although the transition-aware matrix is effective in practice, its estimation is not strictly identifiable without additional assumptions. Therefore, in complex real-world scenarios, it should be understood as a progressively refined and practically useful estimator, rather than an exact recovery of the true label-transition structure. Therefore, in the following research tasks, we will further investigate clearer bias detection mechanisms (e.g., confidence thresholding, robust neighbor selection, or multi-view feature aggregation). Meanwhile, inspired by  \cite{su2025dica}, we will further implement cross-modal semantic alignment in DDMP to guide the disambiguation process better. Although DDMP is developed for partial label learning, the proposed iterative diffusion-based disambiguation mechanism may also be relevant to broader recognition tasks under ambiguous or weak supervision. In particular, Xing et al.~\cite{xing2025locality} studied dense audio-visual event localization by modeling locality-aware cross-modal correspondences to suppress irrelevant signals, which, although different from PLL, shares a similar motivation of reducing misleading information through reliable correspondence modeling. This connection suggests that DDMP may have potential applicability beyond label-space ambiguity resolution, extending to correspondence-oriented or weakly supervised recognition problems such as group activity recognition~\cite{yan2020higcin}, compositional action recognition~\cite{yan2023progressive}, few-shot action recognition~\cite{qu2025mvp,qu2026spatio}, and dense audio-visual event localization~\cite{xing2025locality}.

\subsection{Convergence proof of DDMP} \label{Proof4}

In this section, we provide a theoretical explanation of why DDMP helps disambiguate candidate labels. This paper intends to interpret the DDMP algorithm as an \emph{expectation-maximization}(EM) algorithm. The EM algorithm can be used to deduce how to model and update the ambiguous labels through the transition-aware matrix, thereby improving the classification performance of the model. According to Eq.($\ref{8}$) and Eq.($\ref{10}$), we can further simplify the $\mathcal{L}_{0}$ loss in Eq.($\ref{8}$) as,

\begin{equation} \label{B1}
	\begin{aligned}
		\mathcal{L}_{0}&=-\log p\left(\mathbf{S}_{0} \mid \mathbf{S}_{1}, \mathbf{x}, \mathbf{T}, f_{\phi},\theta \right)
	\end{aligned}
\end{equation}

Introducing the unobserved ground-truth label vector $\mathbf{y}$, the log likelihood of the complete data is,

\begin{equation} \label{B2}
	\begin{aligned}
		\mathcal{L}_{0}&=-\log p\left(\mathbf{S}_{0}, \mathbf{y}\mid \mathbf{S}_{1}, \mathbf{x}, \mathbf{T}, f_{\phi}, \theta \right)  =-\log p\left(\mathbf{S}_{0}, \mathbf{y}\mid \mathbf{x}, \mathbf{T},\theta \right)
	\end{aligned}
\end{equation}

Using joint probability decomposition, we can get,

\begin{equation} \label{B3}
	\begin{aligned}
		p\left(\mathbf{S}_{0}, \mathbf{y}\mid \mathbf{x}, \mathbf{T}, \theta \right) = p\left(\mathbf{S}_{0}\mid \mathbf{y}, \mathbf{T} \right)p\left(\mathbf{y}\mid \mathbf{x}, \theta \right)
	\end{aligned}
\end{equation}

Therefore, the Eq.($\ref{B2}$) can be written as,

\begin{equation} \label{B4}
	\begin{aligned}
		\mathcal{L}_{0}&=-\log p\left(\mathbf{S}_{0}\mid \mathbf{y}, \mathbf{T} \right) -\log p\left(\mathbf{y}\mid \mathbf{x}, \theta \right)
	\end{aligned}
\end{equation}

Then, we show that the DDMP implicitly maximizes the likelihood as follows,

\textbf{E-Step:} In the E-step, we compute the posterior probability $p\left(\mathbf{y}\mid \mathbf{S}_{0}, \mathbf{x}, \mathbf{T}, \theta \right)$ as the distribution of the unobserved ground-truth label $\mathbf{y}$.

\begin{equation} \label{B5}
	\begin{aligned}
		&p\left(\mathbf{y} = y_{c}\mid \mathbf{S}_{0}, \mathbf{x}, \mathbf{T}, \theta \right) = \\& \frac{p\left( \mathbf{S}_{0} \mid \mathbf{y} = y_{c}, \mathbf{T} \right) P\left( \mathbf{y} = y_{c} \mid \mathbf{x}, \theta \right)}{\sum_{i=1} p\left( \mathbf{S}_{0} \mid \mathbf{y} = y_{i}, \mathbf{T} \right)P\left( \mathbf{y} = y_{i} \mid \mathbf{x}, \theta \right)}
	\end{aligned}
\end{equation}

\noindent where Eq.($\ref{B5}$) is the posterior probability that the ground-truth label is $y_{c}$. Therefore, according to Eq.($\ref{12}$), the candidate labels can be smoothly disambiguated to calculate a more stable posterior probability $\mathbf{S}$.

\textbf{M-Step:} In the M-step, the model parameters $\theta$ and the transition-aware matrix $\mathbf{T}$ are updated to maximize the expected log-likelihood calculated in the E step. When $\mathbf{T}$ is fixed and the model parameters $\theta$ are updated,

\begin{equation} \label{B6}
	\begin{aligned}
		\mathcal{L}_{\theta}&= -\mathbf{S} \log p\left(\mathbf{y}\mid \mathbf{x}, \theta \right)
	\end{aligned}
\end{equation}

This is equivalent to weighted maximum likelihood estimation. When model parameters $\theta$ is fixed and the transition-aware matrix $\mathbf{T}$ are updated,

\begin{equation} \label{B7}
	\begin{aligned}
		\mathcal{L}_{\mathbf{T}}&= -\mathbf{S} \log p\left(\mathbf{S}_{0}\mid \mathbf{y}, \mathbf{T} \right)
	\end{aligned}
\end{equation}

By using the Lagrange multiplier method, constraining $\mathbf{T}_{ij}>0$ and $\sum_{j=1}^{Q}\mathbf{T}_{ij} = 1$, the final update formula is:

\begin{equation} \label{B8}
	\begin{aligned}
		\mathbf{T}_{ij}^{e}=\frac{\sum_{i=1}^{N} \mathbb{I}(y_{j} \in S_{i})\mathbf{S}_{ij}}{\sum_{i=1}^{N} \mathbf{S}_{ij}}
	\end{aligned}
\end{equation}

\noindent where $\mathbb{I}(\cdot)$ is an indicator function. The essence of the M-step is to minimize the classification loss. Therefore, during the entire training process of DDMP, the E-step estimates the possibility (i.e., posterior distribution) that each instance belongs to a potential ground-truth label, which provides a weighted basis for parameter optimization in the M-step. In the M-step, the update of the transition-aware matrix $\mathbf{T}$ is adjusted by the weighted distribution of the instances to ensure that each item of $\mathbf{T}$ satisfies the constraints at the same time. The transition-aware matrix $\mathbf{T}$ is used to characterize how the ground-truth label is perturbed by ambiguous labels, thereby generating disambiguation labels. When these two components perform well, the entire training process converges. By iteratively updating $\mathbf{T}$, gradually approaching the most realistic label distribution.

\subsection{The forward process derivation and proof of DDMP} \label{Proof1}


The forward diffusion process, where the latent variable has a non-zero mean distribution $\mathbf{S}_{T} \sim  \mathcal{N}\left(f_{\phi}(\mathbf{x}), \mathbf{I}\right)$, can be expressed in a closed-form representation,

\begin{equation} \label{A1}
	\begin{aligned}
		&q\left(\mathbf{S}_{t} \mid \mathbf{S}_{t-1}, f_{\phi}(\mathbf{x})\right)= \\
		&\mathcal{N}\left(\mathbf{S}_{t} ; \sqrt{1-\beta_{t}} \mathbf{S}_{t-1}+\left(1-\sqrt{1-\beta_{t}}\right) f_{\phi}(\mathbf{x}), \beta_{t} \mathbf{I}\right)
	\end{aligned}
\end{equation}

\noindent where $f_{\phi}(\mathbf{x})$ is a mean estimator set by the encoder. Let $\alpha_{t}=1-\beta_{t}$, then we have,

\begin{equation} \label{A2}
	\begin{aligned}
		&\mathbf{S}_{t-1} = \\ &\sqrt{\alpha_{t-1}}\mathbf{S}_{t-2} +(1-\sqrt{\alpha_{t-1}})f_{\phi}(\mathbf{x}) +(\sqrt{1-\alpha_{t-1}})\boldsymbol{\epsilon}_{t-1}
	\end{aligned}
\end{equation}

\begin{equation} \label{A3}
	\begin{aligned}
		\mathbf{S}_{t} = \sqrt{\alpha_{t}}\mathbf{S}_{t-1} +(1-\sqrt{\alpha_{t}})f_{\phi}(\mathbf{x}) +(\sqrt{1-\alpha_{t}})\boldsymbol{\epsilon}_{t}
	\end{aligned}
\end{equation}

By substituting $\mathbf{S}_{t-1}$ into Eq.($\ref{A3}$), we can obtain,

\begin{equation} \label{A4}
	\begin{aligned}
		\mathbf{S}_{t}
        &=\sqrt{\alpha_{t}}\Bigl(\sqrt{\alpha_{t-1}} \mathbf{S}_{t-2}+\sqrt{1-\alpha_{t-1}} \boldsymbol{\epsilon}_{t-1}+  \left(1-\sqrt{\alpha_{t-1}}\right) \\ 
        & f_{\phi}(\mathbf{x})\Bigr) + \sqrt{1-\alpha_{t}} \boldsymbol{\epsilon}_{t}+ \left(1-\sqrt{\alpha_{t}}\right)f_{\phi}(\mathbf{x})\\
		&= \sqrt{\alpha_{t} \alpha_{t-1}} \mathbf{S}_{t-2}+\sqrt{\alpha_{t}} \sqrt{1-\alpha_{t-1}} \boldsymbol{\epsilon}_{t-1}+\sqrt{1-\alpha_{t}} \boldsymbol{\epsilon}_{t} \\
        &+\sqrt{\alpha_{t}}\left(1-\sqrt{\alpha_{t-1}}\right) f_{\phi}(\mathbf{x})+\left(1-\sqrt{\alpha_{t}}\right) f_{\phi}(\mathbf{x})
	\end{aligned}
\end{equation}

It is known that $\boldsymbol{\epsilon}_{t-1}$ and $\boldsymbol{\epsilon}_{t}$ conform to $\mathcal{N}\left(\mathbf{0}, \mathbf{I}\right)$ Gaussian distribution, so according to the Gaussian function superposition formula, the above formula can be simplified to,

\begin{equation} \label{A5}
	\begin{aligned}
		\mathbf{S}_{t} = \sqrt{\alpha_{t} \alpha_{t-1}} \mathbf{S}_{t-2}+\left(1-\sqrt{\alpha_{t} \alpha_{t-1}}\right) f_{\phi}(\mathbf{x}) + 	\sqrt{1-\alpha_{t} \alpha_{t-1}}\boldsymbol{\epsilon}
	\end{aligned}
\end{equation}

By mathematical induction, we can obtain the relationship between $\mathbf{S}_{t}$ and $\mathbf{S}_{0}$,

\begin{equation} \label{A6}
	\begin{aligned}
		\mathbf{S}_{t} &= \sqrt{\alpha_{t} \alpha_{t-1} \cdots \alpha_{0}} \mathbf{S}_{0}+\left(1-\sqrt{\alpha_{t} \alpha_{t-1} \cdots \alpha_{0}}\right) f_{\phi}(\mathbf{x}) \\ &+ 	\sqrt{1-\alpha_{t} \alpha_{t-1} \cdots \alpha_{0}}\boldsymbol{\epsilon}
	\end{aligned}
\end{equation}

Let $\bar{\alpha}_{t} = \alpha_{t} \alpha_{t-1} \cdots \alpha_{0} = \prod_{t} \alpha_{t}$,we have

\begin{equation} \label{A7}
	\begin{aligned}
		\mathbf{S}_{t}= \sqrt{\bar{\alpha}_{t}}\mathbf{S}_{0} + (1 - \sqrt{\bar{\alpha}_{t}})f_{\phi}(\mathbf{x}) + \sqrt{1 - \bar{\alpha}_{t}}\boldsymbol{\epsilon}
	\end{aligned}
\end{equation}

\subsection{Proof of the PLL generative perspective} \label{Proof3}

From a generative perspective, the PLL can be expressed as follows:

\begin{equation} \label{C1}
	\begin{aligned}
		p(S|\mathbf{x}) &= \sum_{c=1}^{Q}p(S|y=y_{c})p(y = y_{c}|\mathbf{x})
	\end{aligned}
\end{equation}

We provide a detailed derivation for the Eq.($\ref{C1}$). According to the Bayes’ theorem, the conditional probability $p(S|\mathbf{x})$ can be expressed as:

\begin{equation}  \label{C2}
	\begin{aligned}
		p(S|\mathbf{x})  = \frac{p(S,\mathbf{x})}{p(\mathbf{x})} = \frac{\sum_{c=1}^{Q}p(S|y=y_{c})p(\mathbf{x},y = y_{c})}{p(\mathbf{x})}		
	\end{aligned}
\end{equation}

Here, $p(\mathbf{x},y = y_{c})$ represents the joint probability distribution of instance $\mathbf{x}$ and correct label $y=y_{c}$. Eq.($\ref{C2}$) assumes that $p(S|\mathbf{x}, y) = p(S|y)$, given the correct label $y$, the candidate label set $S$ is independent of the instance $\mathbf{x}$. According to the definition of PLL, if $y_{c} \notin S$, we have $p(S|y=y_{c})=0$. Therefore, we can further simplify this expression.

\begin{equation} \label{C3}
	\begin{aligned}
		p(S|\mathbf{x}) &= \sum_{c=1}^{Q}p(S|y=y_{c})p(y = y_{c}|\mathbf{x}) \\&= \sum_{y_{c} \in S} p(S|y=y_{c})p(y = y_{c}|\mathbf{x})
	\end{aligned}
\end{equation}

\section{Conclusion} \label{section_5}

In this paper, an iterative diffusion disambiguation model, called DDMP, for learning on the partial label dataset is proposed. The highlight of DDMP is that it reformulates the label disambiguation problem from the perspective of generative models, where labels are disambiguated through the reverse denoising process. Specifically, an information complementarity label disambiguation strategy is proposed, which can explore the potential information of instances and candidate labels to reduce the impact of ambiguous labels and negative nearest neighbor instances to construct pseudo-clean labels and be used for downstream disambiguation learning. Furthermore, a transition-aware matrix is introduced to estimate the potential ground-truth labels, which are dynamically updated during the diffusion generation. Meanwhile, theoretical analysis shows that this matrix update can be interpreted from an EM-style algorithm perspective. During training, the ground-truth label is progressively refined, improving the classifier. We evaluate the DDMP on partial label datasets and obtain state-of-the-art performance.


\section*{Acknowledgment}

This work is partly supported by the National Natural Science Foundation of China (Grant Nos.62401309, Nos.62401362, Nos.62576183), the China Postdoctoral Science Foundation (No.2025M771692), the Natural Science Foundation of Shandong Province (No. 2R2024QF119), the Qingdao Natural Science Foundation Project (under Project No.24-4-4-zrjj-89-jch), and open project (No.2024PY030).

\ifCLASSOPTIONcaptionsoff
  \newpage
\fi




\bibliographystyle{IEEEtran}
\bibliography{reference}

\begin{IEEEbiography}
	[{\includegraphics[width=1in,height=1.25in,clip,keepaspectratio]{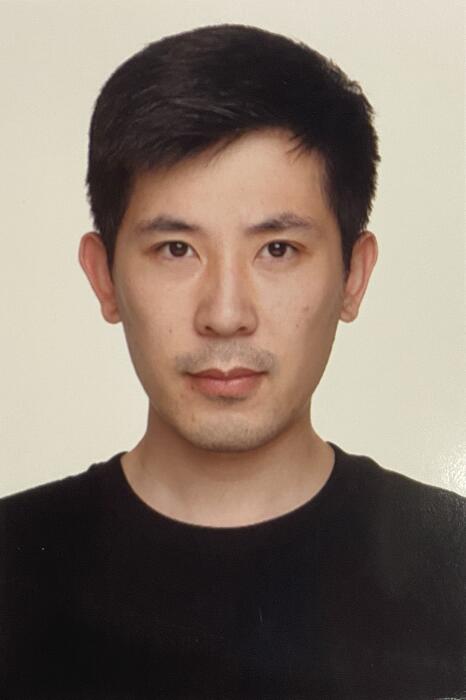}}]{Jinfu Fan} the Ph.D. degree in control theory and control engineering from Tongji University, Shanghai. He is currently an associate professor with the College of Computer Science and Technology, Qingdao University, China. His current research interests include machine learning and computer vision, graph network, especially in learning from partial label data.
\end{IEEEbiography}

\begin{IEEEbiography}
	[{\includegraphics[width=1in,height=1.25in,clip,keepaspectratio]{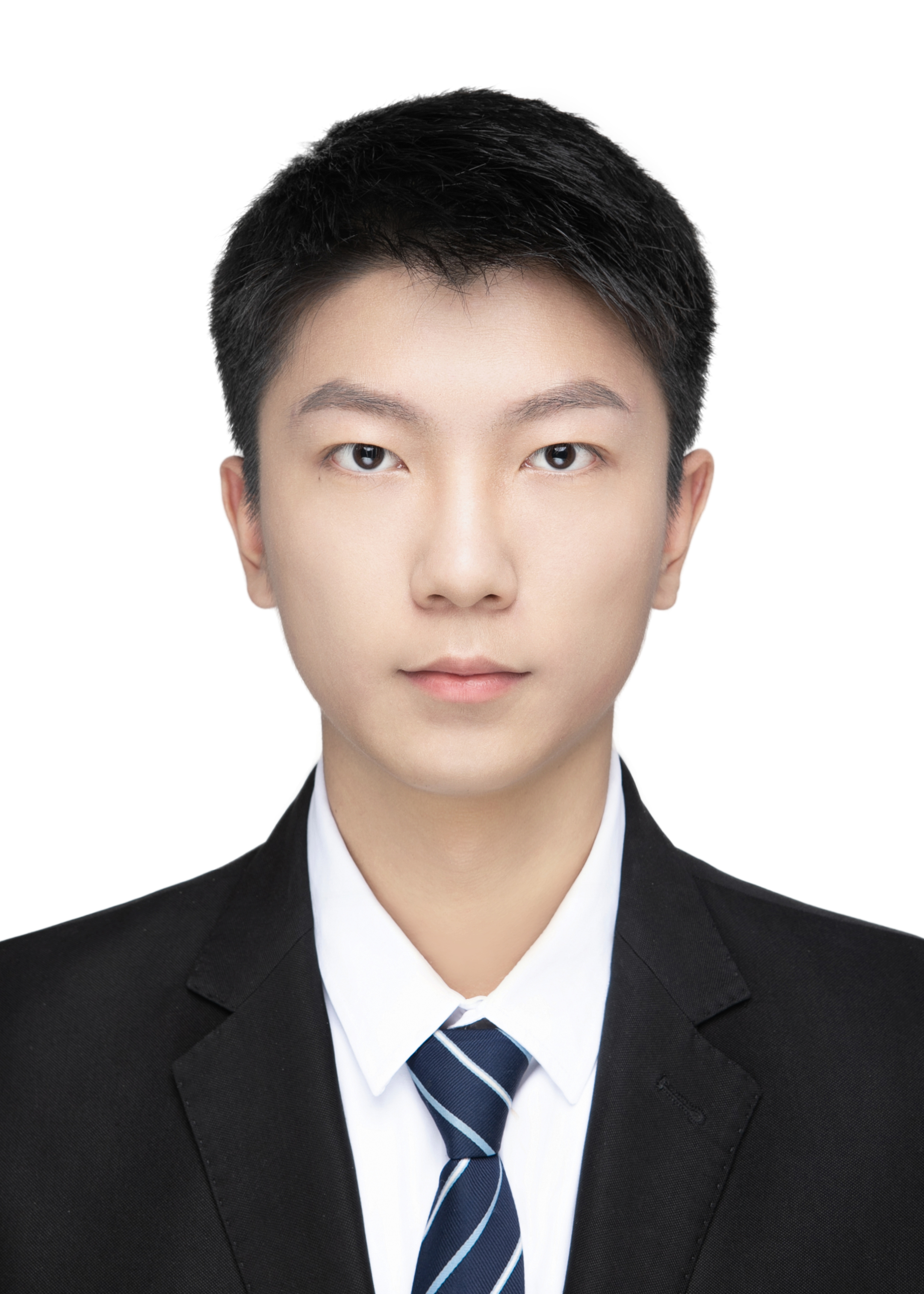}}]{Xiaohui Zhong} received the B.S. degree in software engineering from Qingdao University, Qingdao, China, in 2024, where he is currently pursuing the M.S. degree in software engineering. His current research interests include deep learning methods related to zero-shot learning and partial label learning.
\end{IEEEbiography}

\begin{IEEEbiography}
	[{\includegraphics[width=1in,height=1.25in,clip,keepaspectratio]{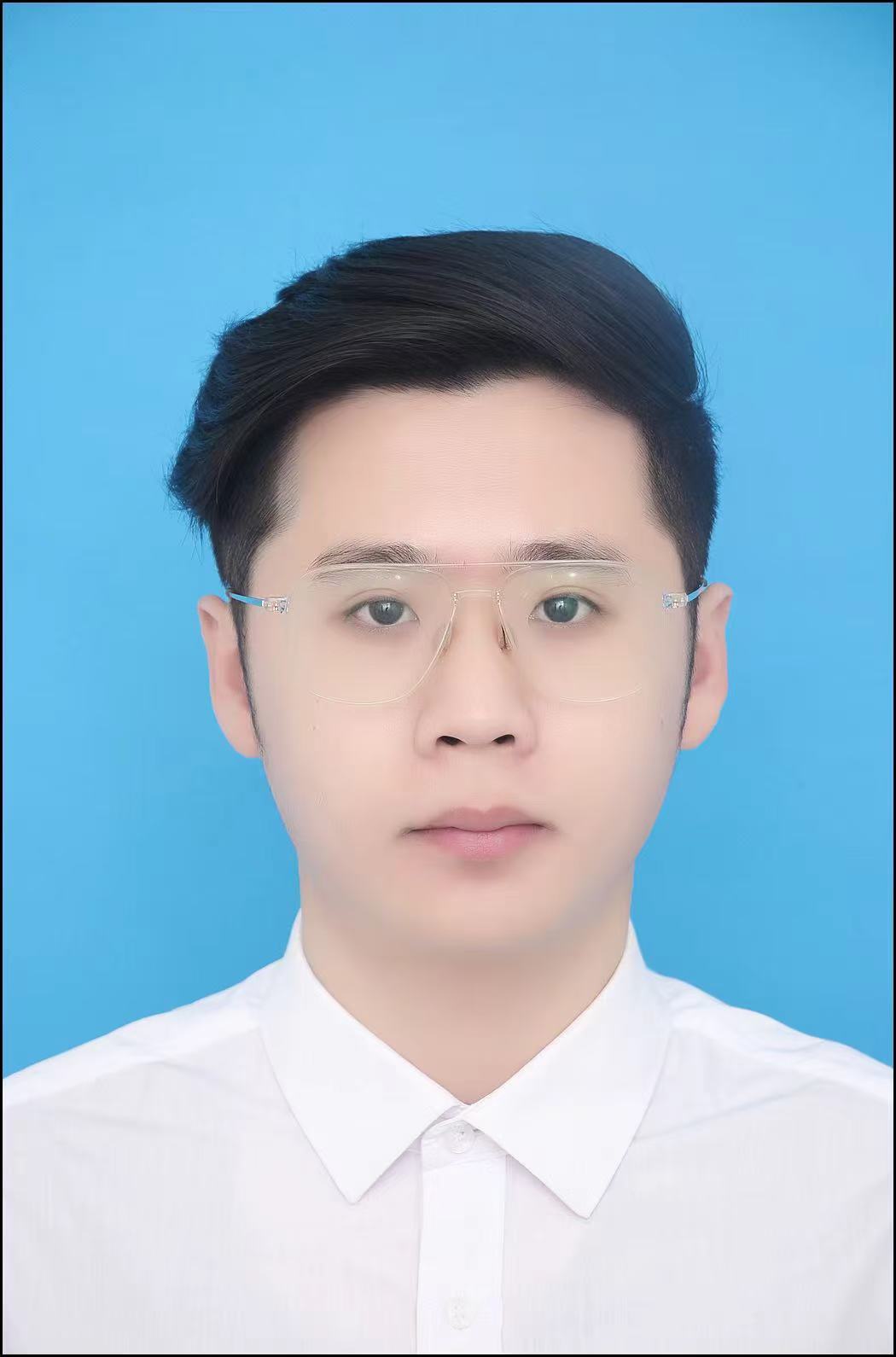}}]{Kangrui Ren} received B.S. degree from Tongji University, China, in 2019, and M.S. degree from University of Southern California, US, in 2020. He is currently pursuing the Ph.D. degree in electronic and information engineering with Tongji University, Shanghai. His current research interests include game theory and machine learning, especially in multi-agents deep reinforcement learning.
\end{IEEEbiography}

\begin{IEEEbiography}
	[{\includegraphics[width=1in,height=1.25in,clip,keepaspectratio]{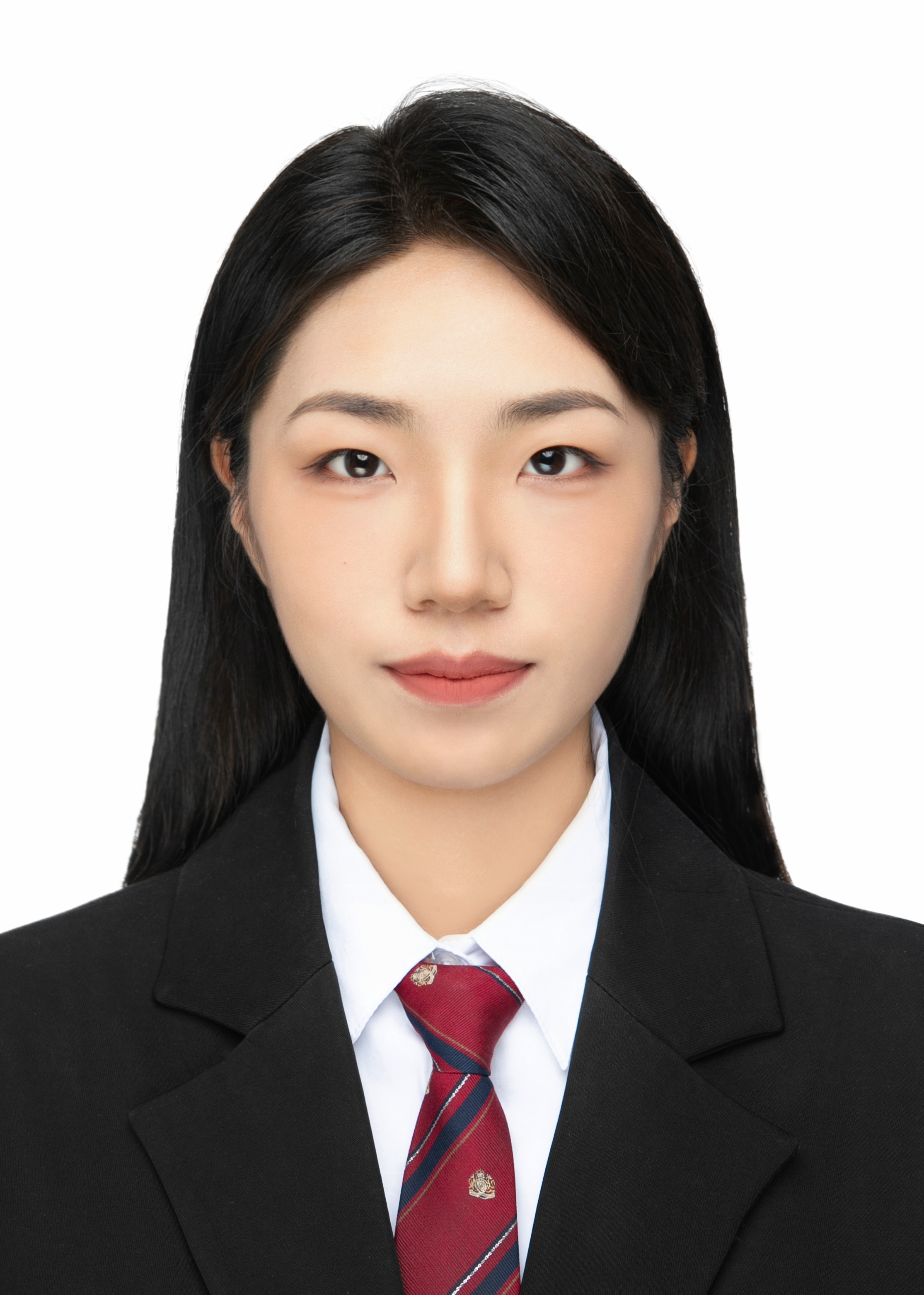}}]{Jiangnan Li} received the B.S. degree in software engineering from Qingdao University, Qingdao, China, in 2024, where she is currently pursuing the M.S. degree in software engineering. Her current research interests include deep learning methods related to zero-shot learning and partial label learning.
\end{IEEEbiography}

\begin{IEEEbiography}
	[{\includegraphics[width=1in,height=1.25in,clip,keepaspectratio]{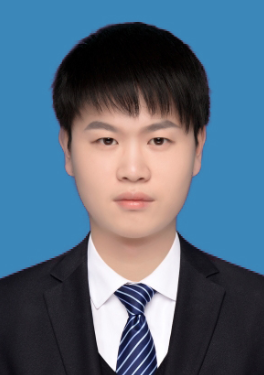}}]{Linqing Huang} is an assistant professor in the School of Computer Science at Shanghai Jiao Tong University (SJTU). He was the visiting scholar in the School of Computing (SoC) at National University of Singapore (NUS) from July 2022 to July 2023. His research interests mainly focus on pattern classification, machine learning, deep learning, especially on cross-domain and low-quality data mining.
\end{IEEEbiography}

\begin{IEEEbiography}
	[{\includegraphics[width=1in,height=1.25in,clip,keepaspectratio]{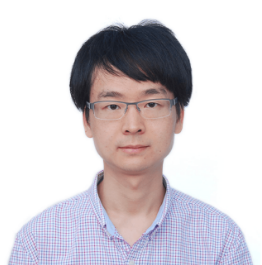}}]{Min Gan} (Senior Member, lEEE) received the B.S. degree in Computer Science and Engineering from Hubei University of Technology, Wuhan, China, in 2004, and the Ph.D. degree in Control Science and Engineering from Central South University, Changsha, China, in 2010. His current research interests include machine learning, inverse problems, and image processing.
\end{IEEEbiography}

\begin{IEEEbiography}
	[{\includegraphics[width=1in,height=1.25in,clip,keepaspectratio]{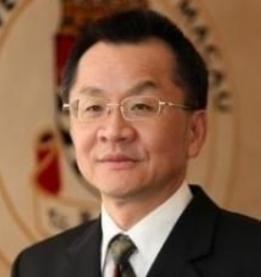}}]{C. L. Philip Chen} (Fellow, lEEE) is the Chair Professor and Dean of the College of Computer Science and Engineering, South China University of Technology. Being a Program Evaluator of the Accreditation Board of Engineering and Technolog Education (ABET) in the U.S., for computer engineering, electrical engineering, and software engineering programs, he successfully architects the University of Macau's Engineering and Computer Science programs receiving accreditations from Washington/Seoul Accord through Hong Kong Institute of Engineers (HKlE), of which is considered as his utmost contribution in engineering/computer Science education for Macau as the former Dean of the Faculty of Science and Technology. Prof. Chen is a Fellow Of IEEE, AAAS.IAPR, CAA, and HKlE; a member of Academia Europaea (AE), European Academy of Sciences and Arts (EASA), and International Academy of Systems and Cybernetics Science (lASCYS). He received lEEE Norbert Wiener Awardin 2018 for his contribution in systems and cybernetics, and machine learnings. He is also a Highly Cited Researcher by Clarivate Analytics in 2018 and 2021.
\end{IEEEbiography}

\end{document}